\documentclass{article}

\usepackage{PRIMEarxiv}

\usepackage[utf8]{inputenc} % allow utf-8 input
\usepackage[T1]{fontenc}    % use 8-bit T1 fonts
\usepackage{hyperref}       % hyperlinks
\usepackage{url}            % simple URL typesetting
 \usepackage{orcidlink}
\usepackage{booktabs}       % professional-quality tables
\usepackage{amsfonts}       % blackboard math symbols
\usepackage{nicefrac}       % compact symbols for 1/2, etc.
\usepackage{microtype}      % microtypography
\usepackage{lipsum}
\usepackage{fancyhdr}       % header
\usepackage{graphicx}       % graphics
\usepackage{xcolor}
\usepackage{enumitem}
\usepackage[numbers]{natbib}
\usepackage{multirow}%
\usepackage{amsmath,amssymb}%
\usepackage{amsthm}%
\usepackage{mathrsfs}%
\usepackage[normalem]{ulem}
\usepackage{subcaption}

\newcommand{\OOD}{OOD~}

\newcommand{\ODD}{ODD~}

\usepackage{todonotes}

\usepackage{lscape}
\usepackage{longtable}
\usepackage{multirow}
\usepackage{tabularx, booktabs}
\usepackage{color, colortbl}
\newcolumntype{R}[1]{>{\raggedright\let\newline\\\arraybackslash\hspace{0pt}}m{#1}}
\definecolor{Gray}{gray}{0.9}

\newcommand{\ILH}[1]{\textcolor{blue}{\textbf{#1}}}

\graphicspath{{media/}}     % organize your images and other figures under media/ folder

\title{Out-of-Distribution Detection for Safety Assurance of AI and Autonomous Systems
\thanks{\textit{\underline{Citation}}: 
\textbf{Authors. Title. Pages.... DOI:000000/11111.}} 
}

\author{
  Victoria J. Hodge\orcidlink{0000-0002-2469-0224}, Colin Paterson\orcidlink{0000-0002-6678-3752}, Ibrahim Habli\orcidlink{0000-0003-2736-8238} \\
  Department of Computer Science \\
  University of York \\
  York, YO10 5GH, UK\\
  \texttt{\{victoria.hodge, colin.paterson, ibrahim.habli\}@york.ac.uk} \\
}

\begin{document}
\maketitle

\begin{abstract}
The operational capabilities and application domains of AI-enabled autonomous systems have expanded significantly in recent years due to advances in robotics and machine learning (ML). Demonstrating the safety of autonomous systems rigorously is critical for their responsible adoption but it is challenging as it requires robust methodologies that can handle novel and uncertain situations throughout the system lifecycle, including detecting out-of-distribution (OoD) data. Thus, OOD detection is receiving increased attention from the research, development and safety engineering communities. This comprehensive review analyses OOD detection techniques within the context of safety assurance for autonomous systems, in particular in safety-critical domains. 
%
%
%It uses this as a platform to develop a system lifecycle for safety assuring autonomous systems which integrates OOD detection throughout the lifecycle stages. 
We begin by defining the relevant concepts, investigating what causes OOD and exploring the factors which make the safety assurance of autonomous systems and OOD detection challenging. 
%We use this to inform the development of our architecture - a system lifecycle for safe autonomous systems where OOD detection is integrated throughout the lifecycle.
%
Our review identifies a range of techniques which can be used throughout the ML development lifecycle and we suggest areas within the lifecycle in which they may be used to support safety assurance arguments. We discuss a number of caveats that system and safety engineers must be aware of when integrating OOD detection into system lifecycles. We conclude by outlining the challenges and future work necessary for the safe development and operation of autonomous systems across a range of domains and applications.
\end{abstract}

% keywords can be removed

\section{Introduction}

Recent AI technological innovations have altered how economies and businesses operate and how people interact with built, social and natural environments. A major innovation is the emergence of autonomous systems, which have the potential to revolutionise our daily lives. One simple definition of an autonomous system is: ``\textit{a system involving software applications, machines, and people, that is able to take actions with little or no human supervision}''\footnote{https://tas.ac.uk/our-definitions/ (accessed 27 May 2025)}. The use of autonomous systems has expanded rapidly to a variety of application domains, in particular safety-critical domains where a failure can cause harm to people, harm to the environment, or substantial economic loss~\citep{bozzano2010design}. Autonomous systems for safety-critical domains include autonomous road vehicles; robots including medical robots, assistive robots, manufacturing robots and delivery robots; aerial vehicles; marine vehicles; underwater vehicles;  wireless sensor networks used for monitoring; and components within systems such as an autopilot system in an aircraft (see \citep{zhang2017current} and section \ref{sec:ooddetection}).

Developing systems that operate in safety-critical and open-world domains remains complex, necessitating the adoption of rigorous and often expensive through-life safety assurance processes to ensure safety standards and regulations are met and that the system is free from unacceptable risks. Traditional safety assurance  is highly conservative, requires the bulk of safety knowledge to be known in advance, and has limited generalisability and scalability \citep{leveson2023certification}. ``\textit{The specific safety challenges of autonomous systems and the technologies that enable autonomy are not adequately addressed by current safety management practices and standards}''~\citep{scsc_as}. System developers often assume independent and identically distributed (IID) training, testing and run-time data that are all drawn from the same distribution known as the ``closed-world assumption''. The transition of AI-enabled autonomous systems from controlled (closed-world) to open-world environments changes the development focus from optimising and guaranteeing against known operational parameters to effectively detecting and mitigating novel inputs, unexpected situations and uncertainty.

The behaviour of autonomous systems alters during their lifetime~\citep{burton2020MindTheGaps,watson2005autonomous}, e.g. due to faults, resource variability, evolving user needs, cyber-attacks.   Furthermore, the deployment environment may be underspecified at design time, or even unknown, and the generalisation of design-time verification to run-time safety is problematic which all increase the likelihood of unknown scenarios.  Thus, autonomous systems lack well-understood safety guarantees and this can lead to unpredictability and ultimately hazardous failures~\citep{hsu2024safety}. 

The objectives of OOD detection is to identify invalid or anomalous inputs, identify uncertainty in the ML model and identify unsafe outputs (for example, unsafe predictions or plans) that could lead to unpredictable errors and compromise the overall safety of the system. Identification may then allow for process modification at design-time or the application of mitigating actions are run-time to compensate for errors occurs due to these inputs.

OOD detection and monitoring is of value in many domains and applications.\citet{yang2024} provide a comprehensive review of general OOD, and \citet{salehi2021} provide a more general review of related topics. \citet{rahman2021} survey run-time monitoring of machine learning for robotic perception which covers OOD detection. Other general AI surveys on anomaly detection~\citep{chandola2009anomaly}, novelty detection~\citep{markou2003novelty,pimentel2014}, open set recognition~\citep{geng2020recent} or outlier detection~\citep{hodge2004survey,hodge2014outlier} cover some aspects of OOD detection and are relevant. However, these reviews and surveys do not focus on safety assurance and the system lifecycle.

Whilst researchers, developers and safety engineers are focusing much effort on OOD detection throughout the system lifecycle, until now, no one has systematically considered how it is planned, developed and applied in the context of safety engineering. In this paper, we provide the first comprehensive and overarching review of OOD for safety assurance, and the safety assurance of OOD detection for autonomous systems. We bring a safety lens to the whole lifecycle of autonomous systems operating in safety-critical and uncertain domains and how to use OOD detection to
support the construction of a compelling safety case \citep{habli2025big,paterson2025safety}. 
For autonomous systems that operate in safety-critical applications, the software itself can contribute to hazards, and its development must be safety-assured. OOD detection can function as an acceptance criterion for software development as discussed in section \ref{sec:MLMV} and \citep{hodge2025agile}.

The rest of the paper is structured as follows: In Section~\ref{sec:developingAS} we provide an overview of the development lifecycle for AI-enabled autonomous systems which is used throughout the paper to structure the results of our survey. Section~\ref{sec:ooddetection} then provides an overview of OOD detection. Having covered this background material, we then outline the ways in which OOD may be used throughout the Autonomous Systems Lifecycle in Section~\ref{sec:usingOOD}. Section~\ref{sec:discussion} then discusses some of the issues identified by our work before moving onto conclusions, challenges and potential avenues for future work in Section~\ref{sec:conclusion}.

%\textcolor{red}{ The rest of the paper is structured as a follows, in section \ref{sec:background}, we define an autonomous system, describe their system development lifecycles using a generic framework, discuss safety assurance, how we demonstrate that an autonomous system will meet its safety requirements, standards and regulations when operating throughout its life, and how safety assurance brings an additional focus on the operational domain of the autonomous system which must be tightly defined.  Section \ref{sec:ooddetection} defines out-of-distribution across a range of learning paradigms, identifies the causes of out-of-distribution data (distribution shifts), briefly outlines how to prove safety, and identifies \OOD application domains across ML paradigms. 
%
%We develop our lifecycle for safety assuring autonomous systems using \OOD detection - how \OOD detection can support and provide evidence for safety assurance from specification, through data development, model learning and verification, to run-time operations and monitoring in section \ref{sec:stages}. We integrate existing work on \OOD detection into the stages of our lifecycle in section \ref{sec:algos}. The discussion in section \ref{sec:discussion} pinpoints a number of caveats when combining \OOD and safety assurance, including safety assuring both the ML component and the system it sits inside. Our review concludes in section \ref{sec:conclusion} with the challenges for safety assurance and \OOD in autonomous systems and the future directions for their research, development and engineering.}

\section{Developing autonomous systems for safety-critical applications\label{sec:developingAS}}

Integrating machine learning into autonomous systems has enhanced their capabilities and broadened their applications \citep{porter2025insyte}. However, many autonomous systems operate in safety-critical domains such as transportation, healthcare or factories. This requires us to include robust mechanisms into their development to handle the inherent uncertainties associated with machine learning models and these safety-critical environments. Safety assurance during the design, implementation and operations is a rigorous process for developing, communicating and challenging the necessary evidence and communicating that sufficient confidence is achieved in the existing and employed risk management measures. It builds in robustness to failures and the ability to handle unfamiliar situations by enabling the system to respond in a way that reduces the risk of accidents, ultimately contributing to safer and more resilient autonomous operation \citep{burton2024resilience, hollnagel2006resilience}.

%In this section we introduce: autonomous systems and their levels of autonomy which influences how we safety assure them; we describe the lifecycle for developing and deploying safe autonomous systems; and then discuss safety assuring these systems during development and once deployed.

\label{sec:autonomousSystems}

In this paper, we consider safety-critical autonomous systems to be those systems that use ML models in the perception and decision-making pipeline of a cyber-physical system, where these models have the capacity to influence hazards and therefore cause harm as defined in~\citet{mcdermid2024safety}.

In this paper, we use the Sense, Understand, Decide and Act (SUDA) model of autonomous systems~\citep{mcdermid2024safety}. The model works as follows: First, sensors capture information about the world, encoding the operating environment for use in the understanding and decision stages. Then, perception components in the understanding stage transform these data into a set of related concepts that the system can reason over. Planning components in the decision block then use world models to predict possible futures and identify the optimal course of action. Finally, low-level controllers on the autonomous system platform execute these plans.

\begin{figure}
\includegraphics[width=\linewidth]{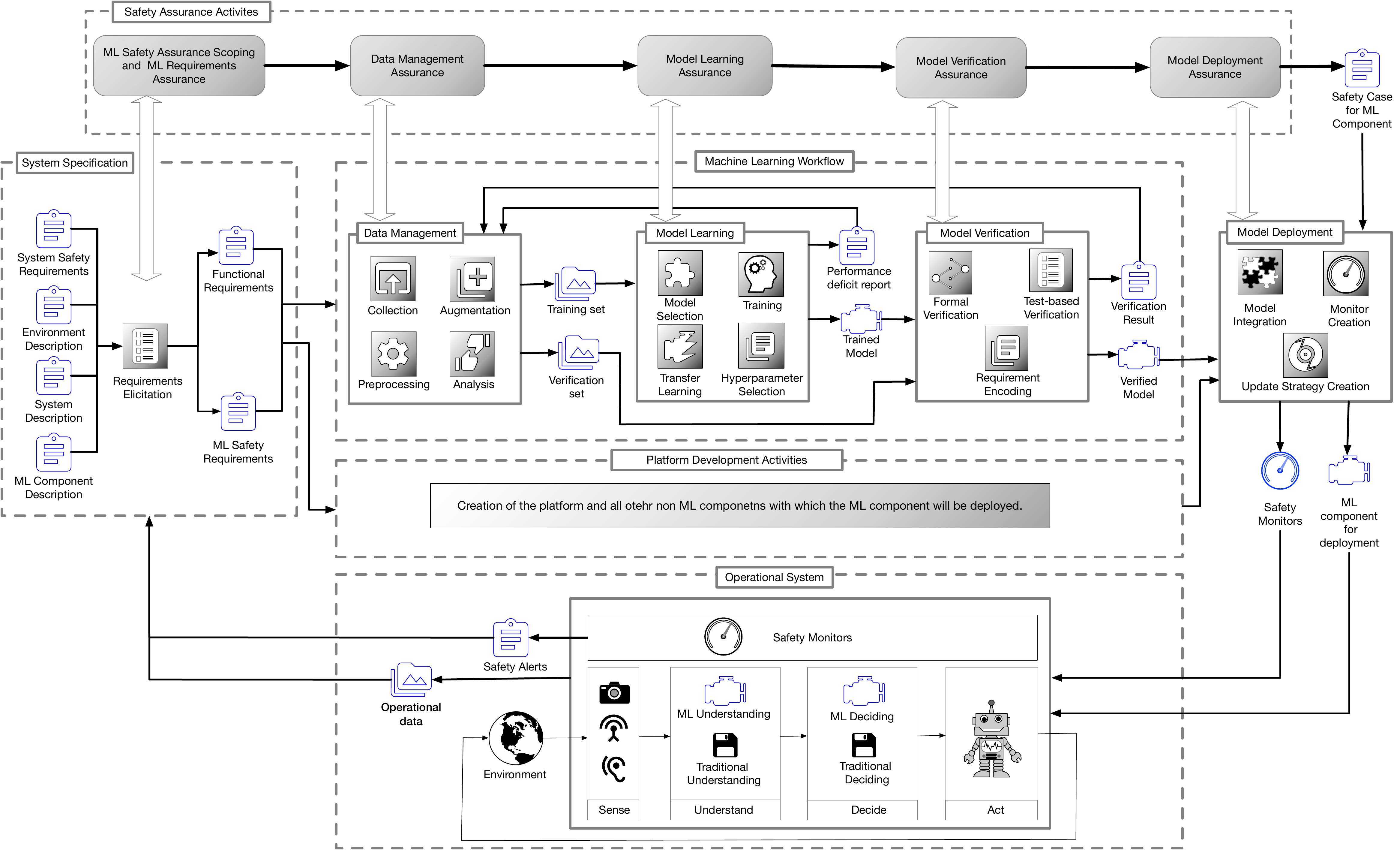}
\caption{The ML lifecycle showing the integration of ML development activities with safety assurance activities, platform engineering activities and run-time operation. Adapted from~\citep{ashmore2021assuring,amlas2021} \label{fig:ML-Process-Diagram}}
\end{figure}

Developing these autonomous systems requires combining ML development, traditional software development, mechanical engineering, and systems safety engineering.
The integration of these activities is shown in Figure~\ref{fig:ML-Process-Diagram} which provides an overview of an autonomous system's development lifecycle adapted from \citep{ashmore2021assuring,amlas2021,hodge2025agile}.

Deciding when the system  is acceptably safe is not straightforward, however, and the process for deciding this varies widely across domains~\citep{dey2021}, required safety levels and applicable safety standards. 
To allow us to decide when the system is safe enough, we need to derive a set of safety requirements, that proactively address assessed and perceived safety risks, and to show that these requirements have been met.  

The most widely used approach to do this is the safety case (or assurance case)~\citep{kelly2004}. A safety case comprises documented evidence and a structured chain of arguments. Together these describe the use case, system design description, environment description and system safety requirements among others as shown in Fig.~\ref{fig:ML-Process-Diagram}. For each ML component in the autonomous system (such as a pedestrian detection component or a lane-keeping component), we take the system level safety requirements and derive component-level safety requirements by analysing the system to identify the role that component may play in contributing to safety hazards \citep{paterson2025safety}. Thus, each component has its own set of allocated safety requirements. We verify and validate the fulfilment of the requirements for our specified operational environment, subject to any stated assumptions regarding the system and its domain. The safety case must then be approved before deployment and maintained throughout the system's operational life \citep{buysse2025safe}. 

For the types of safety critical autonomous systems we are concerned with, the environment changes throughout the lifetime of the deployed system. It is therefore imperative that during run-time operations, the system continues to monitor the fulfilment of the requirements and logs any evidence of violations encountered which would invalidate the safety case. The safety case must, therefore, include not only evidence that that system was developed appropriately but that it will continue to operate safely throughout the lifetime of the system deployment.

The development process begins with the system specification stage which takes a set of documents including the safety requirements, environment description, system description, and ML component description. The environment description and system description together define the operational design domain (ODD) and form the basis under which many of the assumptions underpinning system safety will be made. When the system finds itself outside of this ODD, guarantees of safety can not be assumed to hold. Understanding when this occurs, however, is not simple in complex environments and requires the use of monitoring strategies using techniques such as out-of-distribution detection. The set of documents are then used to derive the system's functional requirements as well as the ML safety requirements for the ML component to be developed. Activities in this stage include identifying how the ML component might contribute to system hazards during operation. The processes undertaken at this stage and artefacts generated feed into the safety assurance pipeline, establishing a safety argument where generated evidence supports the claim that component requirements will satisfy system safety requirements~\citep{amlas2021}.

After deriving the ML-component requirements, we proceed to the ML workflow section of the lifecycle. In this phase, we conduct data management activities to create training and validation datasets that are sufficient to encode the requirements from the previous stage. Our safety assurance argument must address the set of desiderata outlined in~\citep{ashmore2021assuring}: completeness, robustness, accuracy, and balance. We generate two separate datasets to ensure independence in the verification of models used in safety-critical domains; only the training set will be available during model learning.

The model learning stage uses the training data to create a model that meets the component's functional and safety requirements. The assurance process demonstrates the appropriateness of the development approach by examining the model-type selection, parameter derivation, and process execution \citep{hodge2025agile}. The validity of this safety argument relies heavily on both the development team's testing procedures and a well-documented development process with justified design decisions.

Before deployment, model verification must be conducted by independent testers \citep{hodge2025agile}. Testing includes samples designed to evaluate the model's performance limits, as well as formal tests where possible to verify that the model's properties meet specifications. These test results serve as evidence in the safety argument, demonstrating that the model has fulfilled its allocated safety requirements. If the model fails to meet these requirements, the development team must return to earlier stages in the process (either to gather additional data or to redefine the ML-component requirements) where appropriate mitigation strategies can be identified.

Once the model has passed all verification tests, it is deployed alongside the ML components in the operational system as a run-time monitor. In Figure~\ref{fig:ML-Process-Diagram}, we demonstrate this system using the previously described SUDA process, which incorporates ML components in both the \textit{Understand} and \textit{Decide} stages.

In parallel with ML component development, traditional systems engineering practices are undertaken to develop the platform where the ML component will be deployed. The information from these activities, along with the developed ML model, flows into the model deployment stage of the lifecycle.

The model deployment stage involves both component integration and the development of monitoring and update strategies. Monitoring ensures that design-time assumptions remain valid throughout the system's operational lifetime, which are  crucial for safety-critical systems operating in complex, open-world environments. The update strategy specifies how to respond to monitor alerts, defining immediate actions, assessing alert impact, and determining steps to rectify identified issues~\citep{picardi2023transfer}.

Autonomous systems are taking on roles traditionally performed by humans or highly automated systems in well-controlled environments. This shift to autonomous systems led to the development of level of autonomy. The most common framework defines five levels of automation \citep{SAE2021}, ranging from no automation (level 0) to full automation with unlimited operational scope (level 5). A richer framework for the classification of autonomous systems is presented in~\citep{porter2025insyte} which distinguishes between 8 dimensions of autonomy including the complexity of the operating environment and the operational independence of the system with respect to human oversight. In this paper, we are concerned with those systems where the levels of autonomy increase across these dimensions.
%Levels 0–2 require human oversight, while levels 3–5 operate under machine supervision. Most systems discussed in the paper operate at levels 3 or 4 within specified domains. 
Indeed, as autonomy levels increase, deployment environments become less predictable, making it more likely that systems will encounter unknown, unstructured, or dynamic situations. These systems must sense and understand unexpected events and changing conditions while executing their tasks. Consequently, higher autonomy levels demand stronger safety assurance, with out-of-distribution detection serving as a crucial component for managing novelty, uncertainty, and change.

\section{Out-of-Distribution (OOD) Detection an Overview}\label{sec:ooddetection}

Having discussed the operational design domain (ODD), we now have a definition of, and assumptions about, the bounds of operation and, by implication, we can define those things which fall outside of these bounds. Where safety claims are made they are conditional on the context within which the systems will operate. As such, claims are made with respect to the defined ODD. For the safe operation of the autonomous system, it is therefore important to understand when the system moves outside of this domain \citep{hawkins2022guidance}. Knowing when these violations occur is the job of out-of distribution detection (OOD detection).

\begin{figure}[h]
\centering
    \includegraphics[height=3.5cm]{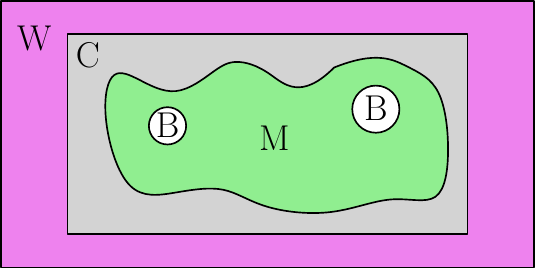}
\caption{The autonomous system's operational environment is the ODD (C). Region $M$ is ID for the ML model. However, there are regions of uncertainty: the grey area of C outside $M$ and the regions labelled B within $M$~\citep{burton2023addressing} caused by insufficient training data or erroneous data for region C, and, insufficient learning  or erroneous learning for B. Regions B and the grey area of C are thus  OOD w.r.t. $M$ but inside the ODD. The pink area outside C is OOD w.r.t. the ODD.  For level 5 autonomy the ODD covers the real world so C=W.\label{fig:BCDworld} }
\end{figure}

Figure~\ref{fig:BCDworld} shows an abstract representation of the real-world where W is the full set of data which could possibly be experienced by the system. Within this space the ODD (region C) is defined as a sub-space of the real-world for which we need to provide assurance for the safe operation of the system. When using AI, we create a model  of the ODD (M) which will be an approximation of the ODD. This approximation means that part of the ODD (grey areas) will not be adequately covered by the model. Conceptually, we assume that these types of errors are due to challenges with collecting sufficient data to ensure coverage of the problem space. Knowing that we are operating in this region will be essential to the safe operation of the system.

Furthermore, there will we regions of the model (labelled B) where errors occur due to limitations of the model arising from the model structure or the training process used to construct the model.

Classifying areas in this way allows us to consider samples which are outside the ODD (pink): samples which are problematic due to issues with collecting the data (grey) and those for which model construction is a problem (white). Using a perception component from an autonomous vehicle (AV), a traffic sign detector, as our example, a Chinese traffic sign for an AV developed for the European market would be outside the ODD and in the pink area. A rare traffic sign only found in a limited number of training images may not be learned by the model so is in the grey area of C. Setting our model training hyper-parameters to achieve the highest metric(s) score for the training data has resulted in the model not learning a particular traffic sign so this is reflected by region B (a gap in the model).

%We note that for many autonomous systems, ML is not confided to a single role within the system, but instead requires the coordination of multiple ML models as shown in Fig. \ref{fig:ASworld}. As such, data that is \OOD for one model may be within distribution for another.

\if 0
\begin{figure}[h]
\centering
    \includegraphics[height=3.5cm]{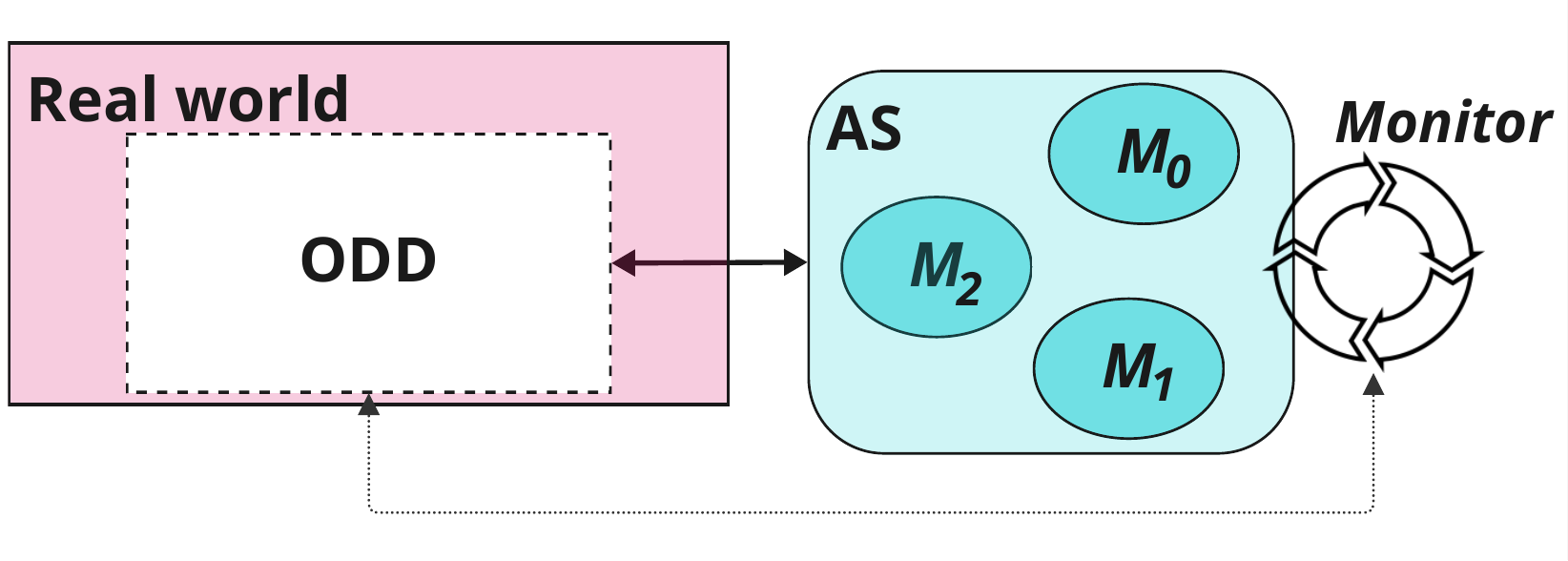}
\caption{An autonomous system is a system of systems that operates in a bounded ODD and will contain a number of ML models ($M_0, M_1, M_2$,...) which must be analysed and monitored throughout their lifecycle.\label{fig:ASworld} }
\end{figure}

\fi

OOD is a commonly used term in the literature which encompasses a number of  techniques and applications including open set recognition and outlier, novelty and anomaly detection. As proposed in~\citet{yang2024}, we adopt a ``generalized OOD'' view which subsumes all of the above categories as they are all relevant for keeping autonomous systems safe across the lifecycle. 
We do however, subdivide OOD by lifecycle stages and the ML model paradigms unsupervised, supervised, semi-supervised and reinforcement learning as this allows us to demonstrate the value of OOD techniques throughout the ML development lifecycle. OOD detection must operate at different levels of abstraction: distribution of data, interaction between agents, and environmental contexts and dynamics. We consider how OOD data arise next.

\subsection{Distribution Shift\label{sec:DistributionShift}}

\begin{figure}
\centering
\begin{subfigure}[b]{0.48\textwidth}
\centering
\includegraphics[width=0.6\linewidth]{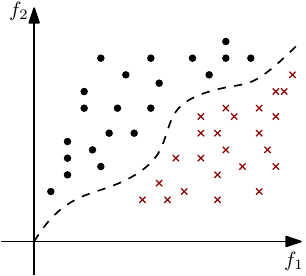}
\caption{Original Distribution\label{fig:shifts-original}}
\end{subfigure}
\begin{subfigure}[b]{0.48\textwidth}
\centering
\includegraphics[width=0.6\linewidth]{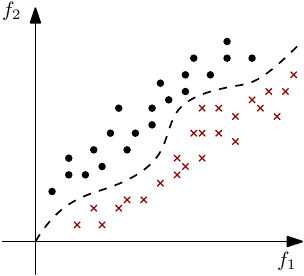}
\caption{Covariate shift\label{fig:shifts-covariate}}
\end{subfigure}
\begin{subfigure}[b]{0.48\textwidth}
\centering
\includegraphics[width=0.6\linewidth]{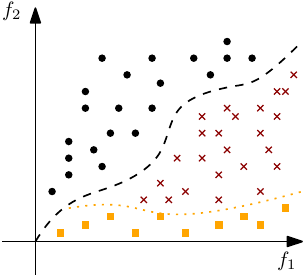}
\caption{Semantic shift: Label\label{fig:shifts-label}}
\end{subfigure}
\begin{subfigure}[b]{0.48\textwidth}
\centering
\includegraphics[width=0.6\linewidth]{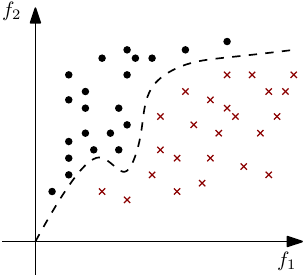}
\caption{Semantic shift: Concept\label{fig:shifts-concept}}
\end{subfigure}
\caption{Effects of distributional shift on a supervised classification problem black circle, red crosses and orange boxes are classes defined on two features ($f_1$ and $f_2$). The dashed line represents ground truth class separation boundaries. \label{fig:shifts}}
\end{figure}

The statistical distribution of data is one way to describe the data we use in the development of our models and to describe the data we see during operation. Unfortunately, these distributions are not static and, where there is a shift between distributions, we can no longer rely on the assumptions made in the construction of our models. As such, it is only through an analysis of that data that we will be able to identify data samples, and trends within sets of data samples, which indicate that the system is operating outside the defined operational domain. In this section, we discuss the nature of this distributional shift in data.
In order to introduce the concepts of distribution shift, we consider a supervised classification problem as illustrated in Figure~\ref{fig:shifts}. Figure~\ref{fig:shifts-original}  shows a simple two-class data problem, black circles and red crosses. The dashed line then indicates the ground truth class-separation boundary of the two classes and the position of each sample reflects the joint probability distribution $P(x,y)$ over the space X $\times$ Y where the covariate space X=\{\{$f_1$, $f_2$\}\} and the label space Y$=$\{black-circle,red-cross\}.

\if 0
\begin{figure}[h]
\centering
\includegraphics[width=\linewidth]{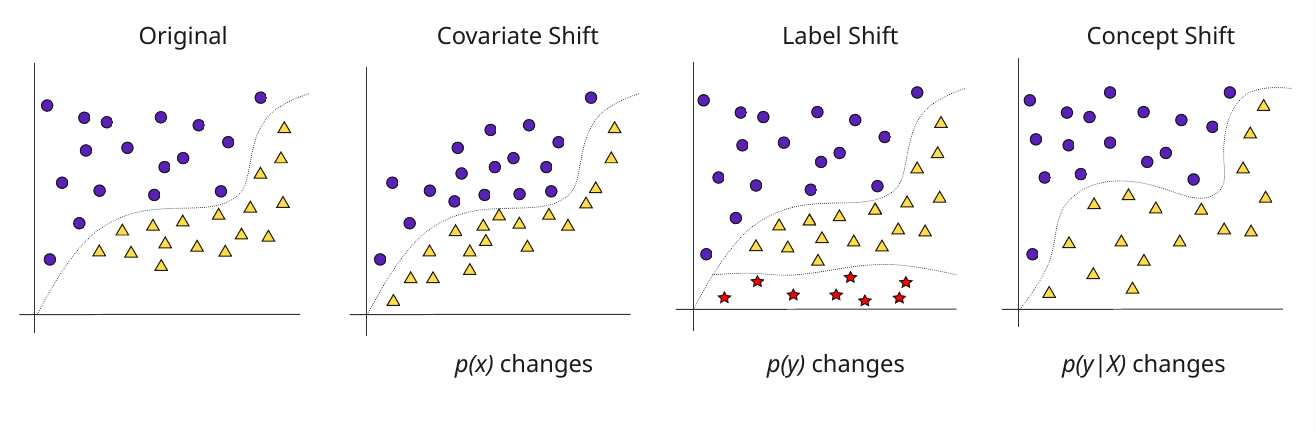}

\caption{3 types of distributional shift as applied to a supervised classification problem with two classes.\label{fig:shifts}}

%\caption{Data distributions for a 2 class problem. The covariate shift moves the data points, i.e., the input data $p(x)$ changes while $p(y | X)$ remains constant. There are two types of semantic shift: concept drift where $p(x)$ does not change but the labelling $p(y | X)$ does and label shift where the label set $p(y)$ changes, but for a given label, the input distribution $p(x | y)$ remains the same~\citep{dmlsbook2022}. \label{fig:shifts} }
\end{figure}
\fi

\textbf{Covariate
shifts} describe how the distribution of the samples across the feature space changes over time. Figure~\ref{fig:shifts-covariate} shows that the boundary remains the same, but the distribution of samples for both classes change such that the data are closer to the boundary. One example of covariate shift is where a vision model trained with images collected during the day is 
deployed at night. Other causes can be found in~\citep{fang2023,hodge2023sensors,koopman2018toward,zhao2024}. Here the features associated with contrast and brightness will lead to a distributional (covariate) shift.

A covariate shift can be described mathematically as \begin{itemize}
\item[]  the input probability density function of the data $p(x)$ changes while  the probability distribution of labels associated with samples $p(y | x)$ remains
constant (as per~\citep{dmlsbook2022})\end{itemize}

Covariate shifts can arise due to a number of factors. 
\begin{enumerate}
\item System faults in the hardware or software of the platform into which the ML components are embedded. For example, sensors such as cameras producing blurred images; sensor mountings degrading and causing increased vibration or; a software module to calculate the ambient temperature assumes Celsius when the system assumes Fahrenheit.

\item Changes in the environmental conditions within which the autonomous system is operating. For example, the amount of rain or fog encountered is greater than the maximum anticipated at design time.

\item Sampling bias. Obtaining balanced data in open-contexts is challenging especially for safety-critical systems where some data may be too expensive or too dangerous to collect. Some areas of the distribution may be over-represented in the training and test data while others may be under-represented. 

\item Switching operating domains. Whilst every effort is made to define an ODD, the reality is that systems may  intentionally be used outside of these domains. For example, a car designed for use on US roads may be imported to the UK where the type of vehicle, weather conditions and even road layout are different to that expected at design time.
\end{enumerate}

\textbf{Semantic shifts} describe how the distribution of labels change over time. For a label shift (Figure~\ref{fig:shifts-label}  the boundary changes due to  new class being added (shown by the orange squares). A concept shift (Figure~\ref{fig:shifts-concept}) arises when the definition of a label changes as new knowledge becomes available and this new knowledge moves the boundary. Some examples can be found in~\citep{islam2023}.

These forms of shift can be described mathematically as:
\begin{itemize}
\item[] a \textbf{label shift} is where the label set $p(y)$ changes, but for a given label, the input distribution $p(x | y)$ remains the same,
\item[] a \textbf{concept shift} is when $p(x)$ does not change but the labelling $p(y | X)$ does.
\end{itemize} Semantic shifts represent a knowledge gap that needs to be identified. They arise due to a range of factors including:
\begin{enumerate}
\item Emergent properties and interactions not accounted for during design and development, for example, the orientation of objects presented at design time is not distributed in the same way as objects encountered at run-time due to a change in the angle of the camera causing a semantic gap~\citep{burton2020MindTheGaps}).

\item Sampling bias where some labels were under-represented or missing from train and test data, as a particular class was too expensive, too dangerous or simply occurred too infrequently to be collected.

\item Agent-centric shifts where the function of an autonomous system changes, for example, in the medical domain, an autonomous robot for cleaning a hospital will encounter a different distribution from a medical delivery robot in the same hospital. The operational domain is the same but the function within it changes. 
\end{enumerate}

Some operational changes of autonomous systems can result in covariate or semantic shifts.
\begin{enumerate}

\item Interaction-centric shifts occur when an autonomous system interacting with different agents, for example, a robot that changes from interacting with medical staff to interacting with patients in the same hospital. This is a non-functional change leading to both a covariate and semantic shift where the operating domain may change. 

\item A contextual shift changes the context of the environment and the autonomous system for example, an autonomous car using a tunnel under a river when  the bridge it planned to use was closed. This could produce covariate and/or semantic shifts. The operating domain is different and needs to be covered in the ODD description.
\end{enumerate}

\subsection{Filters and wrappers\label{sec:FiltersAndWrappers}}

Within this paper, we classify our OOD algorithms into two categories, filters and wrappers. The former is a technique in which the algorithm is integrated with the model, whilst the later is ``wrapped" around the model

A \textbf{filter} uses features of the ML model to determine OOD so the OOD detector is part of the ML model as in equation \ref{eq:filter}. 
\begin{equation} \label{eq:filter}M_{OOD} \subset M \end{equation}
The main advantage of a filter is efficiency as only one algorithm is required. The disadvantage is that
the functional performance of the OOD detector is tightly
coupled with the ML component. 
%\item 

A \textbf{wrapper} uses a secondary algorithm $M_{OOD}$ to wrap around the ML model as in equation \ref{eq:wrapper} to identify OOD data.\begin{equation} \label{eq:wrapper}M_{OOD}(M) \end{equation}The wrapper provides independent OOD predictions that can be tailored to the task (including detecting ID data that cause incorrect inferences (region B in Fig.~\ref{fig:BCDworld})). However, wrappers are often model-specific and there is a trade-off between the functional
and non-functional performance of the ML model and the wrapper~\citep{yuhas2023co} which needs to be considered from a safety assurance perspective. 

In the following sections, $M_{OOD}$ is part of $M$ for filter approaches and a separate model for wrapper approaches. All approaches draw $D_{train}$ the training data from the underlying distribution $P_{train}$. The model $M_{OOD}$ may be trained using a range of ML learning paradigms as discussed next.

\subsection{ML Paradigms for OOD Detection\label{sec:MLParadigms}}

Another way in which we distinguish OOD techniques is by considering the ML paradigm used. Here we consider unsupervised, supervised, semi-supervised and reinforcement learning approaches.

\if 0
i) those OOD techniques which apply $M_{OOD}$ to
provide insights into data without labelling (unsupervised), 
ii) those where  $M_{OOD}$ learns from labelled data (supervised) OOD methods, 
 iii) semi-supervised approaches which require partially labelled datasets to train $M_{OOD}$ and 
iv) reinforcement learning which learns using trial and error.

As noted in the Introduction, there is a range of applications for cyber-physical autonomous systems. However, the range of applications where authors have investigated using OOD in conjunction with safety assurance is much reduced.  Many of the papers in the literature that mention safe autonomous systems use standard image data sets for the analyses and focus on object recognition, segmentation or classification without specifying an intended application domain. 
A handful of OOD review papers detail application domains, including OOD in the robotics domain generally \citep{sinha2023systemlevelviewoutofdistributiondata}, OOD for safe robot missions \citep{nandakumar2024} and OOD for robot perception  \citep{rahman2021}.  \citet{wang2024} survey OOD in autonomous vehicles (AVs), \citet{zhao2024} review OOD for 
analysing AV sensors and \citet{ferreira2020} develop a benchmark for detecting AV camera and LiDAR anomalies. \citet{zeller2024TowardAS} develop a through-life safety framework including OOD detection for safe driverless trains. For each of the four learning paradigms, we provide a list of application domains from the literature.

\fi
\subsection{Unsupervised Approaches}

Unsupervised learning requires no prior knowledge of the data,  using $M_{OOD}$ to provide insights into data without labelling. It can discover hidden patterns or data groupings. Over a period of time, the dataset used for training ($D_{train}$) is susceptible to \textbf{covariate shifts} where the distribution $p(x)$  changes so $D_{train}$ no longer represents the data distribution. These \textbf{distribution shifts}  can have safety implications so we would like to be able to identify this.  Unsupervised models provide a mechanism for such detection. 

ODD filters can estimate the density of the region of the distribution where the data point lies or cluster the data effectively drawing a boundary around the ID data (region M in Fig.\ref{fig:BCDworld}) and can then pinpoint remote data points which are flagged as OOD using distance or density calculations. 

Wrappers, by contrast, use a metric to calculate the distance between the new input data and the training data space.

For these unsupervised techniques we consider $D_{train}$ as an  unlabelled set. It is clustered into $K$ clusters $\{cluster_j\}^K_{j=1}$ as in equation \ref{eq:cluster}.
\begin{equation}\label{eq:cluster}
    f(X) \to C \text{ on } D_{train} \text{ where } D_{train} = \{ x_i\}^N_{i=1} \text{ and } C = \{cluster_j\}^K_{j=1}
\end{equation}
For OOD detection, $cluster_j$ will contain at most a few data points if it is OOD.

Unsupervised learning has few hyper-parameters and the algorithms are explainable for providing evidence for safety assurance but they often do not scale well (for large or high dimensional data). They often make assumptions about the underlying data distribution which may not hold and this also makes them sensitive to noise.

\if 0
\subsubsection{Applications}
For autonomous vehicles (AVs), the applications of OOD detection are multifaceted, ranging from perception to planning. Unsupervised OOD detection can help to assure safety in highly dynamic and uncertain environments including AV perception to identify novel driving scenes (from onboard camera images) \citep{ranjbar2022}, fault detection and isolation during AV localisation \citep{mori2019adaptive} and AV trajectory prediction \citep{farid23a} to prevent downstream planning failures.  

In robotics, OOD detection is vital for ensuring both the safety of the robot and the safety of people and property in its environment. Unsupervised OOD detection helps assure robot navigation using uncertainty maps \citep{verdoja2019deep}, to detect when a robotic arm used for grasping objects is operating in a different environment from its training environment \citep{farid22a} and when a the distances calculated from images produced by cameras in a robotic arm for robotic-guided microsurgery \citep{jungo2023unsupervised} are unreliable. 

Other applications of unsupervised OOD detection include detecting OOD telemetry data from spacecraft \citep{hundman2018detecting} and for network intrusion detection \citep{Aslansefat2020} which combines safety and security.

\fi

\subsection{Supervised Approaches}

Supervised OOD algorithms require us to pre-label data as either within distribution (ID)  or out of distribution (OOD). Similar to~\citep{sinha2023systemlevelviewoutofdistributiondata,yang2024}, we consider that supervised OOD detectors learn a model $M_{OOD}$ of the distribution as in Fig.~\ref{fig:BCDworld} and equation \ref{eq:super} where $\forall i, y_i \in \{ID , OOD\}$:
\begin{equation}\label{eq:super}
   M_{OOD} : X \to Y \text{ on } D_{train} \text{ where } D_{train} = \{ (x_i,y_i)\}^N_{i=1}
\end{equation}

For supervised (labelled data), $D_{train}$ has distribution $P_{train}(x,y)$ where $x \in X$ denotes the covariate and $y \in Y$ denotes the output label. 

During training the verification set ($D_{test}$) is used for parameter tuning to optimise the model. It is also the unseen test set used to evaluate the model after it is trained. 

%When the model is deployed to operation the input data are the inference data. $D_{in}$ are analogous to $D_{test}$, i.e., $D_{in}$ are drawn from $P_{in}$ with density $p_{in}(x,y)$.  In this paper, we conflate $D_{test}$ and $D_{in}$ and use $D_{test}$  for simplicity, $D_{test}$ and $D_{in}$ are both inputs to the trained model that generate an inference, one during training and testing, and one during deployment.

For supervised learning, distribution shift is observed when the 
verification data ($D_{test}$) or operational data ($D_{in}$)  are sampled from a distribution that is different from $P_{train}$. Hence, $D_{test}$ is OOD with respect to $D_{train}$ which is IID.

Supervised approaches tend to outperform unsupervised or semi-supervised as ground-truth (data with labels) are available. However, supervised learning requires complete, relevant, accurate, and fully-labelled training data to work effectively, i.e., no OOD data points in ID training data and good coverage of $P_{train}$ to prevent over-fitting, inconsistencies and gaps in the model~\citep{hodge2018evaluation}.

\subsection{Semi-Supervised Approaches}

Semi-supervised approaches use a small set of labels or implicit labels (ID only or OOD only) and effectively draw a boundary around the in-distribution~\citep{hodge2011outlier}. OOD detectors learn a model $M_{OOD}$ of the distribution using either equation \ref{eq:super} but with only a few $\{ (x_i,y_i)\}$ OR equation \ref{eq:super} but $\forall i, y_i \in \{ID\}$ or $\forall i, y_i \in \{OoD\}$, i.e., there is an implicit label as the data are drawn from only the ID or OOD distribution. $D_{train}$ has density $p_{train}(x,y)$ where $x \in X$ denotes the covariate and $y \in Y$. The distribution may undergo \textbf{covariate shift}s, and \textbf{semantic shifts} of the given or implicit labels. As we often do not know what is OOD a priori and may not even be able to anticipate what OOD is, being able to draw a boundary around ID can help identify anomalies without prior knowledge. They include reconstruction-based filter or wrapper approaches that calculate a reconstruction error between the inference and the actual observed value. Domain-based filter techniques learn ID training data by defining a boundary around the ID class and are often used in safety-critical applications to calculate a boundary (as shown in Fig.~\ref{fig:BCDworld}) rather than calculating or estimating the entire data distribution. 

We include foundation models (FMs) here as they use self-supervised (unsupervised) learning to train the foundation model but then fine-tune this model to produce $M_{OOD}$ using supervised learning. Self-supervised learning generates implicit labels from unstructured data  to train the model. This combination of self-supervised and supervised fine-tuning equates best to semi-supervised learning.
FMs can be used for data density estimation and data reconstruction to identify OOD data and for high fidelity data generation and data augmentation to improve the training data and thus improve model resilience of $M$, for example ensuring a safety margin and improved OOD detection accuracy in $M_{OOD}$.

Semi-supervised techniques have the flexibility of unlabelled data which they can couple with the accuracy of labelled data as obtaining fully labelled data may be too expensive or complex in real-world applications. However, they require clean and complete training data to work effectively, i.e., no OOD data points in ID training data and good coverage of $P_{train}$ to prevent inconsistencies and gaps in the model. They can over-fit if only a few labelled instances are available. They are sensitive to noise in the labelled data or if there is noise in implicit labels (e.g., a training set assumed to be IID contains anomalies).

\subsection{Reinforcement Learning}
The focus of OOD detection for Reinforcement Learning (RL) used in autonomous systems is identifying unfamiliar situations by detecting OOD inputs and OOD environmental. RL trains a decision process (policy) to make decisions on which action an agent (autonomous system) should take to achieve optimal results. For example, the action an autonomous car should take when the car encounters a pedestrian. For autonomous systems, RL tends to use model-free approaches which do not build a model of their environment (such as modelling the entire road traffic dynamics for an autonomous car) and are thus best-suited to open-world and dynamic environments and tasks. OOD detection for RL assures the agent's safety in open-world environments where unknown events cannot be eliminated or even anticipated (known \textit{a priori}) as we cannot capture the full complexity and variability at specification and development time.  For example, unexpected behaviour by a pedestrian or where an autonomous mobile robot in a warehouse encounters a new type of obstacle. 

In RL there are no labelled data $p(x,y)$ unlike supervised learning as the aim of RL is to find an optimal policy for transitioning between states given an action as in equation \ref{eq:RL}
\begin{equation}\label{eq:RL}
\pi^* : S \to A \text{ where }\pi^* \in \Pi  
\end{equation}
RL uses a Markov Decision Process (MDP) which optimises its policy $\pi$ to maximise its cumulative reward by interacting with its environment according to its current state $s_t$ and the reward $r_t$ it receives for each action $a_t$ taken. The MDP is a tuple $<S,A,P,R>$ where $S$ is the state space, $A$ is the action space, the state transition probability (the next state given the current state and current action taken and based on $P$) is defined by $P(s_{t+1},|s_t, a_t)$ and $R(s_t|a_t)$ is the reward function. 

Determining what is ID and what is OOD for RL is challenging \citep{haider2024,nasvytis2024}. Unlike OOD detection for the other paradigms where OOD relates to data, in RL OOD is with respect to the agent's interactions with its environment. There is no ground
truth or labels. Additionally, a sequence of inputs may be relevant to determining the state of the system $s_{t+1}$. \citet{nasvytis2024} categorise \textbf{sensory shifts} which change the observation but not the environment dynamics (analogous to covariate shift  such as an agent trained during the day-time switching to night-time operation), and \textbf{semantic shifts} which change the environment dynamics (such as changing the coefficient of friction on the tyres) and represent a knowledge gap (epistemic uncertainty). Sensory shift (OOD) is therefore defined with respect to transitions where the probability of occurrence at run-time differs from that observed during development. 

To fully assure RL, we would need to explore every state/action pair and develop a policy that handles them all safely. Safe RL can introduce a \textbf{safety filter}. This safety filter monitors the autonomous system (including OOD monitoring) during training, verification and run-time. The filter can intervene by modifying or overriding the autonomous system's action plan if it identifies a likelihood of harm. A semantic shift can be detected at run-time by OOD detection using a \textbf{safety wrapper} that analyses the consistency of the observed behaviour against the learned behaviour.

\if 0
\subsubsection{Applications}
For general robotics, OOD detection must ensure not only the robot's safety but that of humans interacting with it. RL applications include a safety controller with switching logic that can identify unsafe (OOD) scenarios for 
robot navigation \citep{musau2022using}, and detecting OOD scenarios for a multi-joint robot arm 
with a two-fingered parallel gripper performing pick and place \citep{haider2024}, 

For autonomous vehicles, RL-based OOD detection it can estimate model uncertainty for AV pedestrian detection \citep{lutjens2019}, detect anomalous behaviour during AV emergency braking \citep{gardille}, learn AV driving policy by reverting to and learning from a human expert during OOD situations \citep{huang2024}, and help assure AV car following for adaptive cruise control (platooning) \citep{ElSamadisy2024}.

\fi

\section{Using OOD throughout the Autonomous Systems Lifecycle\label{sec:usingOOD}}

In this section, we will consider the Autonomous Systems Lifecycle as shown in Figure~\ref{fig:ML-Process-Diagram} and the assurance processes to identify the role of OoD and the techniques which may be used to provide support for the safety assurance process. A table summarising this section is provided in Table~\ref{tab:Summary}.

\subsection{System Specification}\label{sec:SysSpec}

The \textit{System Specification} stage of development requires us to make decisions about the expected operational characteristics
of the system and its environment, elicit and analyse the system safety requirements \citep{ashmore2021assuring} and to define the system scope. In defining the requirements and the scope (\ODD) we define the expected bounds on environmental conditions in the domains that we want to assure, for example, levels of rain, maximum wind speed, etc. These decisions, and the defined bounds, then form the basis of defining what constitutes our boundary for in or out of distribution. 

At this stage we also define the requirements for ML components to be created transforming the system safety requirements into a set of ML component safety requirements. The definition of domain bounds will be a key factor in the definition of robustness requirements and it is essential at this point in the lifecycle that these requirements are sufficient to ensure that safety is maintained post-deployment.

%During the ML safety assurance scoping activity in Fig.~\ref{fig:ML-Process-Diagram}, any system requirements related to the ML component are allocated to that component as its safety requirements. Scoping also identifies the best ML algorithms to use, assesses data availability and any biases, evaluates trade-offs (such as inference speed versus precision), and plans deployment. The scope (ODD bounds) will be used later to assess the robustness of the model. Considering these bounds specifically for use in \OOD supports our justification argument for the \ODD. This in turn defines the role of \OoD in development and run-time operations. 

At this point in the development lifecycle, we should define the requirements for OOD detection that will be used in later stages. A system-centric approach will allow us to understand how moving out of distribution will lead to adaptation or mitigating actions, for example, and therefore allow us to define the need for OOD identification.

With this in mind, it is important that  OOD detection requirements are quantifiable and measurable to provide developers with actionable targets. For example we may specify that the OOD detector should achieve a false positive rate of X\% at a true positive rate of Y\% for OOD detection rather than simply targeting ``a robust OOD detector''. 

We note that incorporating OOD detection into the autonomous system fundamentally influences the system architecture design decisions. OOD detection is more than a simple feature addition. OOD detectors must, at the very least provide an ID/OOD classification and ideally  produce ``uncertainty scores''. Detecting OOD data may trigger human intervention, safe fallback mechanisms and potentially data and model updates. This needs architectural designs that support real-time decision routing, robust alerting, and potentially a distributed control model.

\subsection{Data Management}

Data plays a key role in the development of ML components. The training data is, effectively an encoding of the requirements derived in the system specification stage. It is therefore imperative that the data collected is a true representation of the intention encoded in such requirements.

In previous work~\citep{ashmore2021assuring} we have shown that there are four features which should be addressed in assuring the sufficiency of data to meet the safety requirements specified.  With respect to \ODD we might describe these as follows:   \textbf{Complete}: the data collected for training, test and verification covers the intended ODD; \textbf{Relevant}: data which is OOD is not present in the training data; \textbf{Accurate}: the processes of data collecting, preprocessing and augmentation have not introduced unintended distributional shirts and; \textbf{Balanced}: the distribution of data both within and between classes of data are appropriate for the the purpose of model training testing and verification.

In the analysis stage of \textit{Data Management} OOD techniques may be used to assess these features and highlight the need for additional data collection or, indeed, changes to the requirements specification.

We also recognise that, historically, ML models were trained under a ``closed-world assumption'', i.e., that all future input data will be from the training data distribution $P_{train}$. However, the uncertainties inherent in real-world deployment due to system and environmental variation means that many autonomous systems operate in ``open-world settings'' where ``unseen'' data are encountered. This requires an enhanced approach to data management which recognises the limitations of all data sets collected at design-time.

Through a survey of the literature, we identified a number of OOD techniques which can be used in the \textit{Data Management} stage to support the development of safety-critical ML components.

The first set of techniques we describe can be used to \ILH{``Argue the completeness of the data with respect to the ODD"}.

\textbf{Data Augmentations} \citep{tack2020csi} including \textbf{Geometric transformations} \citep{golan2018deep} are a supervised filter technique which can be used on image data to shift training data instances. A detector is then trained on these instances to identify that a transformation is present in the sample. The intuition is that by learning these transformations encourages the learning of features which are useful in novelty detection.

Collecting data which is on the boundaries of the ODD is expensive if not impossible and \textbf{Generative models} provide an approach to tackle this. These may be classified as semi-supervised filters which show promise for accurately modelling complex data distributions, identifying gaps in these data and synthesising data to augment the train, validate and test data. \citet{dionelis2022} use a Generative Adversarial Network (GAN~\citep{goodfellow2020generative}) reconstruction error approach to generate OOD training samples on the boundaries of the ID data distribution (negative data augmentation) with the aim of improving data coverage. The GAN learns the augmented distribution and can then produce an OOD score at inference time.
%
%Pix2Pix (a type of GAN) and Stable Diffusion can generate synthetic datasets for safety-critical detect and avoid systems \citep{lyhs2025bootstrapping}. This is particularly useful within the Data Augmentation activity. Together they can insert new objects into various background scenes and generate new images. This synthesis can address the scarcity of data which is particularly important for safety critical edge cases. 
%
Dream-Box~\citep{IsaacMedina2025Dreambox} uses diffusion models to generate OOD objects directly into the pixel space which can then be used for OOD detector training. The synthesised OOD objects can be viewed which helps humans to understand the OOD detector's failure modes.

More recently \textbf{Foundation models} have been proposed for the generation of rare cases to aid in training OOD detectors. FodFoM \citep{chen2024fodfom} combines multiple foundation models to produce two sets of OOD images for training models. FodFoM combines BLIP-2’s image captioning capability, CLIP’s vision-language knowledge, and Stable Diffusion’s image generation ability to generate the first set of OOD images which are semantically similar to but different from ID images. FodFoM then uses GroundingDINO’s object detection ability to construct background images by blurring foreground ID objects in ID images. 

Another area where we can utilise ODD techniques in the data management stage is to \ILH{``Argue the relevance and accuracy of data with respect to the ODD"}. Here the aim is ensure that the data is sufficient to allow the  constructed model to be able to discriminate between ID and OOD samples. These techniques are  applied during the analysis activity to assess the quality of our data sets.

The \textbf{labelling} of the training data needs to be accurate for supervised and semi-supervised techniques. \citet{humblot2024noisy} implement 20 OOD detection algorithms to analyse their ability to discriminate between ID/OOD when trained using unreliable labelling of the training data across different data sets. The authors found a clear reduction in overall OOD detection performance when the algorithms are trained using data with label noise. They show that these effects are not systematic and not easy to predict. They identify that many algorithms are unable to separate incorrectly classified ID samples from OOD samples. The authors recommendation is to introduce uniform
(synthetic) label noise into the training data. 

Other approaches detect the \textbf{label noise} so data with incorrect labels may be rejected or corrected. Detection techniques include using Gaussian mixture models \citep{Li2020DivideMix}, using the neighbourhood coherence of contrastive learning representations \citep{ortego2021multi} or prediction uncertainty evaluations using contrastive learning and two different views of each sample to estimate its ``likelihood'' of being clean or OOD \citep{yao2021jo}. Correction techniques use semi-supervised learning including \citet{li2023DISC} who classify two views of an image using a DNN and threshold the DNN's learning momentum to characterise noisy data which are corrected using regularisation.
% https://openaccess.thecvf.com/content/WACV2023/papers/Albert_Is_Your_Noise_Correction_Noisy_PLS_Robustness_To_Label_Noise_WACV_2023_paper.pdf

Another use for OOD detection is in ensuring consistency between different partitions of the data collected.
 
\textbf{Energy-based models}~\citep{Grathwohl2020Your} is an unsupervised filter method which can learn an ``energy" function which has low scores for ID data and high scores for OOD data.
The authors reinterpret a standard classifier $p(y|x)$ as an energy-based model by estimating the joint distribution (density function) $p(x, y)$ of an energy-based machine from a generative perspective. Whilst this method can detect OOD images, the authors note that energy-based models are difficult to train effectively. 

We can utilise supervised filters to create \textbf{Classification models} which can learn to discriminate ID from OOD data. \citet{Hendrycks2016softmax} test the input data of DNNs during \textit{Data Management} for OOD data by \textbf{model uncertainty estimation} using maximum softmax probability (MSP) of a DNN classifier as a discriminator between ID and OOD. 

A number of semi-supervisied filter techniques exist which can be utilised in this role.

Firstly, we may use \textbf{Rule-based and statistical}  approaches. \citet{koopman2018toward} test domain models against a set of required constraints (safe operational envelope) to assure input data. 
\citet{kaur2024conformal} introduce temporal dependencies into statistical data shift detection for time-series data. Most techniques described here do not consider temporal data and its implications for safety. The authors determine the data's deviation from the ID temporal equi-variance by calculating multiple predictions on sliding data windows (time-series), combining the results using Fisher's statistical method, and determining if the combined value is within pre-determined ID bounds.

Both \textbf{Generative models} and \textbf{Foundation models} also have a role to play here. \textbf{Generative models} can learn the underlying distribution of the ID feature space by reconstructing input data and calculating the \textbf{reconstruction loss}.  \citet{borg2023ergo} calculate the reconstruction error of an autoencoder (AE) using mean-squared error. However, this is less accurate for data on the boundary of the  distribution but can be improved by combining the reconstruction error with Mahalaonobis distance calculated from the mean latent embedding~\citep{denouden2018improving}. \citet{an2015variational} proposed using the reconstruction probability between the input and the reconstructed instance from a variational autoencoder (VAE \citep{Kingma2014}) for identifying OOD data. 
\citet{cai2020} also use VAEs with deep support vector data description (deep SVDD) to learn models to identify new time-series inputs that are OOD relative to the training set. It can perform OOD detection on high-dimensional time-series inputs.

\textbf{Foundation models} can also generate data using textual prompts. NegPrompt \citep{li2024learning} uses  Contrastive Language-Image Pre-training (CLIP \citep{radford2021learning}) to learn negative prompts (negative connotation of class labels) to delineate ID and OOD images. NegPrompt learns using ID data only and can learn negative prompts to generate novel ID classes not seen during training (label shifts). It learns positive prompts to describe ID data first, freezes the learned model and then learns the negative prompts.

\subsection{Model Learning \& Model Verification\label{sec:MLMV}}
\textit{ML Model Learning} is an iterative process in which many models are
created and assessed before changes are made to the hyper-parameters
to refine the model in light of the analysis. During \textit{Model Learning}, we integrate OOD detection components into the ML pipeline to analyse each instance of the ML model $M$ when determining the best version of the model to use. This allows us to assess if the model $M$ covers the data distribution of the ODD ($M \triangleleft$ ODD), whether the system bounds specified in the ODD need revising, and whether $M$ is accurate across the distribution. By analysing the types of OOD data identified, developers can identify gaps in $D_{train}$ and incorporate new data to improve robustness and increase the safety margin.

\textit{Verification and Validation} for safety critical systems is separate from and independent of software testing.  ML \textit{Model Verification} in Fig.~\ref{fig:ML-Process-Diagram}:  uses a parallel environment that mimics the production environment, such as a simulation or mock-up. This step is to ensure that performance and safety properties are met by the ML model and also to validate the intention of the requirements are respected. Whilst the independence of testing may be different, the purpose of OOD techniques across both stages is shared, we therefore consider both stages together in this paper.

%The system safety case which is created in parallel with the ML component, Fig.~\ref{fig:ML-Process-Diagram} will include evidence produced by \OOD techniques throughout these stages. that the system meets its intended safety requirements and meets any relevant safety guidelines, regulations or standards.  The safety case covers the design, and specification of the system, its development and verification and the expected run-time operations which are specified in the ODD and system specification artefacts. 

We note that, at any stage during \textit{Model Learning \& Model Verification}, we may identify the need to collect more data or update the ODD. If this is the case then we can return to the first lifecycle stage \textit{System Specification} and update our requirements and specifications. Indeed, OOD can form part of the acceptance criteria for determining if the autonomous system software fulfils its safety requirements and is ready-for-release~\citep{henriksson2023} by using the frequency and accuracy of OOD detection as metrics.  

OOD detection in these stages can identify regions of the training data where the classification or prediction metrics are below a pre-specified safety threshold. In Fig.~\ref{fig:BCDworld}, $M$ is IID for the ML model. However, there are regions of model uncertainty (labelled B) within $M$~\citep{burton2023addressing} caused by insufficient training data, insufficient learning, erroneous data or erroneous learning. Hence, $M$ has not fully covered its allocated ODD so regions B and the grey areas of C are OOD w.r.t. $M$ but inside the ODD. Thus, OOD detection can determine if the ODD boundary is correct and if this boundary needs systematically expanding.  OOD detection can also identify regions of the ODD where scenarios are not fulfilling their allocated safety requirements or where assumptions do not hold.

Our survey identified a number of OOD techniques which can be applied in this stage of the lifecycle and we have clustered these into sets by role within the stage.

Firstly OOD detection may be used to \ILH{Detect Model learning failures} which can lead to over or under confidence. Since we have now created a model, we can begin to use both wrapper and filter techniques.

Considering supervised wrapper techniques first, we identified a set of \textbf{Statistical Approaches} which can map features from a pre-trained classifier onto class conditional Gaussian distributions using Gaussian discriminant analysis during \textit{Verification}, to produce a confidence score based on the Mahalanobis distance~\citep{lee2018neurips} which can then be used as a threshold. 

\textbf{Entropy-based models} can identify inference errors using post-processing by extracting (pixel-wise) inference uncertainty in images via the entropy measure from the softmax output of a convolutional neural network (CNN) performing semantic segmentation \citep{bruggemann2020}. \citet{Liu_2023_CVPR} developed an entropy-based score function, which can be applied to any pre-trained softmax-based classifier. 
The generalized Entropy score amplifies minor deviations of a predictive distribution
from the ideal one-hot encoding to identify OOD data using thresholding. 

\textbf{Density-based} modelling can outperform distance-based modelling  as it can detect both local OOD data (data that is different from neighbouring data) and global OOD data (outliers with respect to the whole dataset). \citet{luan2021} train monitors for the hidden layers of a DNN using isolation forest (distance-based) or local outlier factor (density-based) algorithms on each layer. These monitor the layers and identify OOD data.  Local outlier factor performs best in the analyses in the paper.

When we consider supervised filter techniques we identified \textbf{Model estimation} approaches which can be used to analyse softmax confidence. \citet{Hendrycks2016softmax} demonstrates this by monitoring the network's maximum softmax probability (MSP). The authors also augment this with an auxiliary (wrapper) decoder which reconstructs the input and improves performance on some datasets. In further work, \citet{hendrycks2019} use maximum logit (MaxLogit) probability as it scales better than MSP.  \citet{liu2020} calculate an energy score using the  denominator of the softmax activation which they demonstrate aligns with the density of inputs an can pinpoint low density regions. 

\textbf{Entropy loss} techniques can alse be used to identify OOD samples. \citet{blei2022identifyingoutofdistributionsamplesrealtime} trained a CNN for object detection using margin entropy (ME) loss to enable the object detector to detect OOD data but maintain object detection performance. The ME loss tackles the CNN's over-confidence on OOD data by using the ME loss to detect when the softmax confidence is high for OOD data - the ME loss is correspondingly high but low for ID data.

Authors have also \textbf{amended the NN's architecture} to overcome the problem of over-confidence in spurious inferences on OOD input data. OpenMax~\citep{bendale2016towards} replaces the softmax layer with a layer that compares the input sample using a probability derived from the NN's penultimate layer. The replacement layer models the training set and correlated classes. \citet{devries2018learning} add a confidence branch at the logits before the softmax layer of a DNN to classify data as ID or OOD.  \citet{sun2021react} adjust the model by attenuating the activation on the penultimate layer units using a threshold and thus reducing over-confidence.

The next set of techniques can be used to \ILH{Detect functional model uncertainty}.    
Functional Model Uncertainty is a concept that falls under the umbrella of OOD~\citep{sinha2023systemlevelviewoutofdistributiondata}. This is represented in Fig.~\ref{fig:BCDworld} as region B where there was a gap in the training data or the model failed to learn a region of the training data correctly. 
This problem can arise when there is scarcity of OOD data during training which limits the diversity and representativeness of OOD samples learned by both $M$ and $M_{OOD}$. This results in functional uncertainty where the model has not fully captured the data distribution.  These models may perform well on the ``known'' OOD data but perform poorly on novel (``unseen'') OOD data. Analysing model predictions and the model's uncertainty can directly improve the robustness and generalisability making the model less susceptible to unseen OOD data. 

Some OOD approaches use OOD data to train the model while others do not~\citep{tian2021exploring}. Both approaches do not have access to $P_{test}$. The former samples the distribution $P_{test}$ and the latter uses a generalised model of $P_{test}$. Training data cannot realistically be large enough to capture all scenarios~\citep{houben2022inspect} so there may be gaps and inconsistencies. OOD detection can identify epistemic uncertainty (semantic shifts), identifying where the model lacks knowledge and aleatoric uncertainty (covariate shifts) stemming from noise, randomness or data ambiguity. However, the autonomous system may need to handle a blurred road sign (covariate shift) differently from semantic shift (a new road sign never seen before) so run-time monitors should be multi-faceted.

There are many OOD techniques which have applicability in this role. A number of unsupervised wrapper techniques have been identified. 
\textbf{Distance-based approaches} such as  
the Laplacian distribution can model uncertainty of object detection distances predicted by a convolutional neural network (CNN). This creates a map of obstacle distance estimates and levels of uncertainty that can be thresholded \citep{verdoja2019deep}.
~
\textbf{Statistical analyses} - student’s t-distribution-based adaptive unscented Kalman filter (T-AUKF) can evaluate the behaviour of sensors for vehicle positioning using the predicted output and its covariance~\citep{mori2019adaptive}. OOD is identified using the correlation between the data generated in the sensor and the prediction. T-AUKF is able to handle noisy data as the sensor's measurement noise is updated adaptively.
\citet{Aslansefat2020} calculate the Kolmogorov-Smirnov distance to monitor the estimated accuracy of an ML component and generate an alert if the deviation exceeds pre-specified boundaries.
~Other authors perform OOD detection using \textbf{self-supervised learning}. ~\citet{hendrycks2019} combine
different self-supervised geometric translation prediction
tasks in one model, using multiple auxiliary heads.
 \citet{Mohseni_Pitale_Yadawa_Wang_2020} train a wide residual network using self-supervised learning which adds nodes to the last layer of the network to act as reject functions for OOD data at inference time. 

We also identified a number of supervised wrapper techniques including \textbf{Model uncertainty estimation}. \citet{averly2023unified} compare: MSP (maximum softmax probability), MLS (maximum logit score), energy score, gradNorm (gradients) and a hybrid post-processor (wrapper) for detecting both covariate and semantic shift in image data. They find no post-processor excels across datasets or models but MSP is the best overall.

Whilst Bayesian Statistical methods can be used to quantify uncertainty they have shown to have problems in scaling for real-world systems. This limitation has been overcome somewhat by authors who approximate the Bayesian posterior by \textbf{ensembling of predictions} from discriminative classifiers trained on IID data~\citep{lakshminarayanan2017simple} or using Monte-Carlo dropout~\citep{gal2016dropout}. We note however that these approximations are not sufficient for most safety-critical run-time applications~\citep{schwaiger2020uncertainty}, only for verification and low safety integrity level (SIL) systems - see \citep{Lohn2020EstimatingTB} for a discussion of SILs.

Where the ML component being considered is an object detector, \textbf{Bayesian safety approaches} have been used to combine fault tree analysis with Bayesian estimation for a risk-aware OOD detection~\citep{yuhas2023co}. 

\citet{han2022} show how an \textbf{ensemble} of multiple DNNs (CNNs, ResNet and inception networks) can be used to detect OOD data during run-time inference, and hence this could be used in the verification stage of the development lifecycle. They use a variety of model architectures and learning mechanisms to ensure diversity and to outperform the individual models. \citet{Wilson_2023_ICCV} perform OOD object detection by extracting object-level feature maps from the layers of a DNN that are most sensitive to OOD data and from object descriptors using `region of interest' pooling on the feature maps. The resultant vectors are concatenated layer-wise and applied to an auxiliary multi-layer perceptron which is trained to generate an OOD score for each image object. The OOD data for training are adversarially-perturbed ID images.

\textbf{Temperature scaling} is a supervised filter approach which proposes a post-processing calibration technique. ODIN \citep{liang2017enhancing} uses input pre-processing (adding small perturbations) and temperature scaling of the scoring function to make the max class probability of the softmax layer of a DNN to identify OOD samples.  The differences of energy scores between ID and OOD allow effective differentiation at run-time and for optimising the network during training.

We also found a number of semi-supervised approaches with Generative models and Density based approaches which can be considered as wrappers.

\textbf{Generative models} are the most common approach we found in this category and assess the likelihood of an input  belonging to a specific data distribution. For verification and run-time monitoring, many authors use autoencoder-based reconstruction approaches. These likelihood-based generative models do not require labelled data and are able to model the input distribution by fitting the generative model to the ID data. The model produces an OOD likelihood score for new data~\citep{ren2019,Serra2020Input}.

\citet{hussain2022deepguard}  introduce an autonomous driving system (ADS) comprising an autoencoder (AE) and time series analysis–based system to prevent safety-critical inconsistent  driving behaviour of the ADS at run-time. It compares the current and reconstructed driving scenarios and predicts potential inconsistent behaviour using a threshold.  \citet{Ruff2020Deep} introduce an end-to-end AE based on the idea that the entropy of the latent distribution for normal data should be lower than the entropy for OOD data. Similar to one-class SVMs, it learns a
minimum hyperplane characterized by a centre and radius so that the sphere contains all training ID instances. They refine this using a few OOD examples to maximise the class separation.   

\citet{feng2021} train a \textbf{variational autoencoder (VAE)} using optical flow information extracted from a time-series window in a video sequence. The OOD score calculates the KL-divergence~\citep{kullback1951information} between the trained VAE and a specified prior. \citet{park2018} combine a VAE and a long short-term memory (LSTM~\citep{hochreiter1997long}) network to fuse time-series signals and reconstruct their expected distribution. It  uses a state-based threshold (which changes with the estimated state of a task execution) to threshold the reconstruction score and determine if it is OOD. 

An \textbf{Adversarial autoencoder} \citep{beggel2020} is a probabilistic network that combines the adversarial training of a generative adversarial network (GAN) with the reconstruction capability of an AE to combine the
reconstruction error with the likelihood in the latent space.  \citet{akcay2019ganomaly} use a GAN with AEs and an adversarial training loss function to perform OOD detection using reconstruction-based distance calculations. 

\textbf{Density-based} approaches can use  multiple kernel learning for one-class classification where the weight for each kernel is assigned locally \citep{gautam2019}. \citet{theissler2017detecting} use a deep support vector data description (one-class classifier) to model the training data as a hypersphere and consider data outside this distribution as OOD.

\textbf{Distance-based} is a semi-supervised filter approach that uses a threshold on an error calculation or a distance measure to identify novelty. \citet{chen2020} use saliency between feature maps of the layers of a trained convolutional neural network (CNN). This identifies the features of the training data space that are most relevant to the prediction. Dissimilar saliency maps, and prediction/saliency maps that do not match the learned model have high prediction error and indicate novel inputs.  OOD detection can use the softmax of scaled cosine similarity on the  NN's output layer \citep{techapanurak2020hyperparameter}. It is hyper-parameter free and can be used with any network by replacing the output layer.

Finally in this stage, we found a number of  \textbf{RL-based Approaches for Functional Uncertainty Detection}. OOD detection  for RL algorithms in autonomous systems that operate during verification and run-time monitoring are a mix of filters and wrappers. The wrapper approaches often use a standard ML algorithm to wrap around the RL algorithm. We include them here as they illustrate how wrappers for RL operate, how they differ from standard wrappers as they learn from aspects of the RL algorithm such as actions, states or policy. Thus we show how they contribute to assuring RL.

For RL-based Filter techniques we identified two approaches: the first is statistical and the second uses a human-in-the-loop.
A \textbf{statistical} sampling-based safety monitor can analyse trajectory planning  \citep{farid23a}. It calculates whether the observed trajectory cost lies within the \textit{p-quantile} of the distribution of the cost predicted using a Markov Decision Process (MDP); and flags the new trajectory as OOD if true.  PAC-Bayes theory \citep{farid22a} provides task-based OOD for RL with statistical guarantees. It learns certified bounds on the RL policy during training to bound the expected performance to the training distribution. At verification and run-time, violating the bounds produces a statistically grounded alert for detecting OOD. 

\citet{huang2024} develop a \textbf{human-in-the-loop RL} (\textbf{HRL}) method to learn a driving policy of an actor-critic deep RL algorithm for an autonomous vehicle were the human supervises and intervenes in the policy learning. \citet{huang2024} ensure continuity for state transitions between the human expert and the agent. This aims to mitigate potential OOD issues if learning relies solely on the human (imitation learning) and does not take into account policy learned from free exploration by the agent.

When we consider RL-based Wrappers we see that \textbf{Generative models} can act as policy guides. The critic in an actor-critic DNN can learn safe policies but the safety filters provide no guarantees so \citet{hsu2024safety} combine deep learning with model predictive monitoring, thus treating the learned controller as an untrusted oracle to guide the safety fallback strategy, which can then be verified at run-time through rollouts to obtain robust safety guarantees.  

Other approaches such as DEXTER \citep{nasvytis2024} use an unsupervised wrapper which extracts features from the time-series of agent inputs (state dimensions) to form the inputs to an OOD detector built from an ensemble of isolation forest models. The ensemble outputs
an OOD score. \citet{prashant2025guaranteeing} use a semi-supervised conditional variational autoencoder (CVAE) to learn the transition dynamics of the training environment and detect deviations for verification and run-time. It compares the transition selected by the policy against the CVAE inference. Similarly, \citet{gardille} learn the dynamics of the agent's current environment using a supervised MLP to learn the RL transition function. They use an adapted student's T-test to compute an OOD score by comparing the output of a supervised model and the current state. They can then threshold the OOD score to identify dynamics that are OOD.

\subsection{Run-time Operation}

\textit{Run-time monitoring} is critical to the safe operation of autonomous systems in open environments which require the system to detect and safely adapt to unknown scenarios which deviate from the anticipated ODD.  Whilst OOD plays a vital role in monitoring, it is only one part of the autonomous system monitoring functionality. Run-time monitors for autonomous systems need to be multi-faceted and monitor: sensor data, contextual data, data flow, the model (e.g., latency), and the model inferences. Run-time monitoring also encompasses monitoring that the safety argumentation is valid. Run-time monitoring should operate at multiple levels of abstraction from the data to models to components up to the system level. In Fig.~\ref{fig:ML-Process-Diagram} this would be the \textit{Safety Monitors} in \textit{Operational System} which were developed during \textit{Model Deployment}.  

OOD detection models provide a continuous process for analysing run-time input data and comparing them against the known (training) data distributions. OOD detection needs to detect OOD data instances and aggregated data distribution shifts. It needs to monitor the data over different time-frames from detecting rapid onset changes, (for example, a lighting failure in a warehouse where a robot is operating) to subtle and slow data shifts, (for example, changing plant growth for an agricultural robot).

Once deployed, each monitoring alert may lead to a different proposed action to address the observed situation. These may include generating an alert which requests a human operator to intervene or triggering a safe fallback strategy with reduced performance.
The operational environment will almost certainly change over time, for example, a new road layout for an autonomous vehicle or new operating legislation for a UAV. OOD detection can trigger a retrain of the ML model ($M$), the OOD detector model $M_{OOD}$ and the associated safety case (including ODD). 

We note that many of the algorithms and techniques described in the previous sections (sections \ref{sec:SysSpec} to \ref{sec:MLMV}) may also be used at run-time but, for brevity, we do not repeat them in this section.

The first role we consider for the use of OOD at run-time is the the \ILH{Detection of anomalies}. These OOD detection algorithms continuously monitor new data inputs to ensure  they are sufficiently similar to the training data distribution used to train the ML model. Anomalous inputs are flagged and appropriate mitigating actions taken, such as triggering a safe fallback routine or alerting a user. The following techniques are all classified as semi-supervised wrappers.

Rule-based techniques can use \textbf{formal verification}. \citet{stahl2021} develop a formal verification approach for run-time monitoring of the trajectory of an autonomous driving system. They allocate safety requirements to the ML component and then develop metrics to monitor whether the trajectory can be safely executed and use rule-based reachable sets to handle dynamic objects during verification.

The \textbf{Density of a Gaussian mixture model-based framework} is used to model the healthy operation of a robot \citep{schnell2020}.  New inputs are scored, by calculating the average log probabilities, which quantifies the level of anomaly in the input data  using its deviation from healthy operation (dense areas).

\textbf{Generative models} have been used to perform lookahead OOD detection by predicting the next state of an autonomous maritime vessel using a simulated prediction in ODDIT \citep{isaku2025digital}. A digital twin built from recurrent neural networks replicates the behaviour of the vessel and estimates its next state. An autoencoder then predicts whether this future state is out-of-distribution.

\textbf{Foundation models} are used in autonomous vehicles to enhance perception, prediction, and planning in dynamic environments. \citet{greer2024towards} show how language-augmented latent representations can be used for safety monitoring of autonomous cars. They use Contrastive Language-Image Pre-training (CLIP)~\citep{radford2021learning} embeddings (data representations) to cluster the dataset so that OOD images can be detected (images whose embeddings are not in a cluster).  They then use language-augmented latent representations for run-time safety monitoring of autonomous cars. It focuses OOD detection efforts on scenarios of interest which can be expressed in natural language such as clear, bright, and open roads. Any deviations should be flagged as OOD inputs. 

The \textbf{Prior-augmented Vision Transformer}, PViT aims to improve the robustness of Vision transformer (ViT) models \citep{dosovitskiy2020image} used in perception components by identifying OOD data using quantification of the divergence between the predicted class logits and the prior logits (prior knowledge) obtained from pre-trained models \citep{zhang2024pvit}. The use of prior knowledge enables interpretability with respect to whether the model is confident or uncertain. This interpretability can help inform a human operator to diagnose issues or design mitigation strategies.

\citet{Cultrera2023} \textbf{combine} \textbf{foundation models} and \textbf{generative models} using convolutional autoencoders (AE) to reconstruct the attention heatmaps generated by a ViT classifier. This improves the accuracy of image reconstruction as the ViT captures fine-grained features and global context and can distinguish visually similar classes while the AE uses the convolutions to learn fine-grained features specific to each class. These perspectives combine to improve OOD detection.

We identified one filter technique, \textbf{Normalising flows} (NFs) represent complex probability distributions with a learnable series of transformations that map from the simple base distribution to a complex target distribution.  For robot perception (object classification), \citet{Feng2023NF} replace the base distribution with a class-conditional resampled base distribution and apply NFs directly on the feature space of the object classifier. They incorporate a post-hoc OOD detector using NFs with flexible, class-conditional base distributions  and output a log-likelihood of OOD.

The next role we consider is in \ILH{Detecting shifts in the data distribution}. Data distribution shift monitoring tracks aggregate changes in the data distribution over time. These changes indicate a systemic shift that may require, at the very least, augmenting the model training data and will probably require full retraining of the model, including new hyper-parameter selection and even new algorithms. For safety assurance, we must also update the ODD and associated safety assurance artefacts. This requires a return to the first stage of the lifecycle - \textit{System Specification}.

Detecting shifts of this type requires the continuous monitoring of input data, the ML model's performance over time (both $M$ and $M_{OOD}$) and model inferences. Distribution shifts can happen suddenly such as a dramatic change in the weather or may be slow and subtle changes. Irrespective of the time scale, distribution shifts will significantly degrade model accuracy. OOD detection algorithms are crucial for analysing whether new data inputs are sufficiently similar to the training data distribution $P_{train}$ and identifying new classes. The following data shift monitors are all \textbf{wrapper} approaches and we first consider approaches to identify \ILH{covariate shifts}.

\textbf{Foundation models} are unsupervised wrappers which can can perform zero-shot covariate shift detection. \citet{heng2025detecting} use vision-language models such as Contrastive Language-Image Pre-training (CLIP)~\citep{radford2021learning}. The authors' approach does not require task-specific training or fine-tuning as models derived from CLIP suffer performance degradation under covariate shifts. These shifts can be detected using Mahalanobis distance to pinpoint distribution shifts over time.

We also identified semi-supervised wrappers for covariate shift. Firstly an uncertainty-aware Bayesian statistical planner \citep{filos2020} for autonomous vehicles using \textbf{ensembles of autoregressive neural density estimators}. It can detect some distribution shifts where the model’s uncertainty is high and recommend a safe course of action by querying the expert driver for feedback. Striveworks (\url{https://www.striveworks.com/}) uses an \textbf{autoencoder} to characterise datasets by computing low-dimensional embeddings (latent space) and applies statistical tests between new data and the latent space, such as Mahalanobis distance for single input OOD detection, and Kolmogorov-Smirnov and Cramér-von Mises tests to compare aggregate data and quantify distributional shifts.
\textbf{Temperature scaling} introduces a temperature parameter to adjust the output of a DNN's softmax function. \citet{hsu2020generalized} extend ODIN \citep{liang2017enhancing} to drift detection by modifying the input data pre-processing of ODIN's DNN to only require ID data.

\textbf{Statistical process control} (SPC) is a supervised wrapper able to identify \ILH{semantic shift}. SPC is a statistical approach that monitors and controls the quality of a process over time and can differentiate expected variation in a process (common cause variation) and other sources of variation. \citet{zamzmi2024out} use geometric metrics such as cosine similarity and Mahalanobis distance to quantify differences in image data. By using SPC charts, they can visualize these metrics and alert the use to any semantic shifts.

\citet{Cultrera2023}'s combined \textbf{foundation model} and \textbf{generative model} approach discussed earlier in this section can identify distribution shifts as well as detecting anomalies. The combination of foundation and generative approaches improves semantic shift detection as it allows analysis from multiple perspectives (global, local, intra and inter class).

Finally in this section we consider techniques to \ILH{Detect novel scenarios and \ODD shifts}.

Novelty detection can identify unseen scenarios not seen during model learning or verification. If detected, these novel scenarios need to be assessed during post-hoc analysis to determine their source and any mitigations required. These scenarios may then be incorporated into future training data. This will progressively enhance the system's competency but requires us to loop back to the \textit{System Specification} stage to update the ODD and associated safety assurance artefacts.

The novelty detection algorithms identified were all \textbf{wrapper} approaches with the majority adopting the semi-supervised learning paradigm.
 
A \textbf{run-time uncertainty measure} can identify when task-salient parts of the image are uncertain. To achieve this \citet{mcallister2019} estimate a measure of the uncertainty of a generative convolutional VAE's prediction using an action-conditioned Bayesian predictive model that maps the projected input states to task-relevant predictions.

\textbf{Generative models} have been used to learn complex data distributions, such as multi-modal human driving behaviour. These generate trajectories of all vehicles in the current scene for prediction and planning in an AV controller \citep{zheng2025diffusionbased}. This synthesis can be aligned with safe driving styles using a classifier. The data synthesis can incorporate OOD data to improve resilience and reduce the need for safe fallback protocols when OOD situations are encountered.

\textbf{Foundation models} can synthesise fallback strategies for autonomous systems at run-time. FORTRESS~\citep{ganai2025real} uses foundation models to generate fallbacks in real-time to prevent OOD failures in autonomous mobile robots. It adaptively synthesises fallbacks using reach-avoid path planning analysis guided by fallback goals and semantic constraints. A new fallback is deemed safe if it is not close to a known failure state. Similarly, \citet{Sinha-RSS-24} develop the  AESOP framework that compares the current situation to previous experience and then the use of foundation models (particularly large language models) for detecting and mitigating out-of-distribution failure modes in robotic systems by identifying OOD and then selecting the best fallback strategy.

Whilst the previous techniques are semi-supervised, \citet{musau2022using} wrap a deep RL controller with switching logic and a safety controller to ensure safety. The switching logic uses a real-time reachability algorithm for short time horizons of states into the future which provides safety guarantees and detects potential unsafe scenarios. It passes control to the safety controller when unsafe scenarios are detected. 

%% Note: dropped continuous learning for now as it felt hard to link to OOD assurance etc. I think this might need more work or even another paper.

\subsection{Other techniques}

During our survey we identified a number of algorithms which are widely applicable throughout the ML development lifecycle and we highlight them here. These are wrapper approaches which develop and auxiliary model for OOD detection.

Various, unsupervised, \textbf{distance-based measures} can analyse the data through-life: for data management, model learning, model verification and run-time monitoring. Training DNNs to model normal data points allows for the identification of OOD data using distance calculations between points (one-class OOD detection). The distance metrics include Euclidean distance~\citep{goyal2020} and Mahalanobis distance~\citep{denouden2018improving,jungo2023unsupervised}.  \citet{sun22d} use non-parametric nearest-neighbour distance so they do not impose any distributional assumption for OOD detection whilst \citet{sikar2024evaluationautonomoussystemsdata} propose three distance metrics (KL divergence (entropy), histogram overlap and Bhattacharyya distance) and empirically derive thresholds to define safe operation limits for data distribution shifts. 

\textbf{Density-based} techniques operate throughout the lifecycle to  estimate the probability density of the ID data space.
\citet{ranjbar2022} use unsupervised feature learning and von Mises-Fisher clustering to detect novelty in camera images for autonomous driving using the clusters. 
\citet{ramakrishna2022efficient} partially disentangle the latent space of a $\beta$-variational autoencoder network ($\beta$-VAE)
tuned using Bayesian optimisation. They learn an approximate mapping between the latent variables and image features by clustering the most informative latent variables and detect OOD from the clusters.
\citet{zong2018deep} use a deep autoencoder (AE) to generate a reconstruction error for each input data point and also use a Gaussian mixture model to perform density estimation OOD detection in the low-dimensional space learned by the AE.

For semi-supervised techniques we identified   \citet{Kirchheim_2024_WACV} who introduce an \textbf{ensemble logical reasoning} approach with three components: a perception system of DNNs, a state updated by the perception system, and the logical reasoning system that checks if the current state is compatible with a set of constraints. The authors note that logic reasoning against constraints can be inflexible and adding a probabilistic component would improve performance. 
\textbf{Foundation model} approaches have been growing in popularity and have been used to analyse data and detect OOD samples throughout the lifecycle. \citet{fort2021exploring} introduce zero-shot learning using pre-trained  transformers such as contrastive language-image pre-trained (CLIP) multi-modal foundation model. The authors demonstrate CLIP's ability to detect OOD images that are near-OOD (close to ID data and harder to distinguish) with over 90\% accuracy just using the names of outlier classes as labels.

Finally, we identified a set of supervised wrappers with open set recognition approaches.
\textbf{Semantic segmentation} classifies the pixels in an image into groups (classes) such as road or car. It can run throughout the lifecycle to find objects in images and classify them. However, many semantic segmentation models operate under a closed-set assumption, where they are only trained to classify objects from a predefined set of classes. Semantic segmentation models produce predictions with high confidence for OOD data as they lack an explicit mechanism for identifying unknown inputs. OOD segmentation in contrast, assigns an OOD score to each pixel to separate ID pixels and classes from OOD elements \citep{shoeb2025out}. This open-world segmentation can identify unexpected and potentially hazardous obstacles.

%\begin{landscape}
\sffamily
\footnotesize
\begin{longtable}{R{2cm}R{3cm}R{1.5cm}R{2.5cm}R{5cm}}
\caption[Some info]{Summary of the OOD techniques identified and areas of application in the safe ML development lifecycle\label{tab:Summary}}\\
\toprule
    \textbf{Stage} & \textbf{Role} & \textbf{Type} & \textbf{ML Paradigm} &  \textbf{Techniques} \\
    \midrule
    \rowcolor{Gray} Data Management & Arguing Completeness & Filter & Supervised  & Geometric Transformations~\citep{golan2018deep} \\
    & & & Semi-Supervised & Generative Models~\citep{dionelis2022, IsaacMedina2025Dreambox}  ,  Foundation Models~\citep{chen2024fodfom}  \\  \cmidrule(l){2-5}
    \rowcolor{Gray} & Arguing Relevance and Accuracy & Filter & Unsupervised & Energy-based models~\citep{Grathwohl2020Your}  \\
    & & & Supervised & Classification Models~\citep{Hendrycks2016softmax}  \\
    \rowcolor{Gray} & & & Semi Supervised & Rule-based and Statistical, Generative models~\citep{borg2023ergo,denouden2018improving,an2015variational,Kingma2014,cai2020} , Foundation models~\citep{li2024learning, radford2021learning}  \\ \midrule
    Model Learning and Verification & Detect Model learning failures & Wrapper & Supervised & Statistical~\citep{lee2018neurips}, Entropy-based~ \citep{Liu_2023_CVPR, bruggemann2020}, Density-based~\citep{luan2021} \\
    \rowcolor{Gray} & & Filter & Supervised & Model Estimation~\citep{Hendrycks2016softmax,hendrycks2019}, Entropy Loss~\citep{blei2022identifyingoutofdistributionsamplesrealtime}, Amended NN Architecture~\citep{bendale2016towards, sun2021react} \\ \cmidrule(l){2-5}
     & Detecting functional uncertainty & Wrapper & Unsupervised & Distance-based~\citep{verdoja2019deep}, Statistical Analysis~\citep{mori2019adaptive,Aslansefat2020}, Self-Supervised Learning~\citep{hendrycks2019,Mohseni_Pitale_Yadawa_Wang_2020}\\
    \rowcolor{Gray}& & & Supervised & Model uncertainty estimation~\citep{averly2023unified}, Ensembling predictions~\citep{lakshminarayanan2017simple,gal2016dropout,schwaiger2020uncertainty,Lohn2020EstimatingTB}, Bayesian safety~~\citep{yuhas2023co}, Ensembling models~\citep{Wilson_2023_ICCV} \\
    & & Filters & Supervised & Temperature scaling~\citep{liang2017enhancing} \\
    \rowcolor{Gray}& & Wrapper & Semi-supervised & Generative models~\citep{ren2019,Serra2020Input,hussain2022deepguard,Ruff2020Deep}., Variational Autoencoder~\citep{feng2021,park2018}, Adversarial Autoencoder~\citep{beggel2020}, Density-based~\citep{gautam2019,theissler2017detecting}. \\
    & & Filter & Semi-supervised & Distance-based~\citep{chen2020,techapanurak2020hyperparameter} \\
    \rowcolor{Gray}& & Filter & Reinforcement Learning & Statistical~ \citep{farid23a, farid22a}, Human-in-the-loop~\citep{huang2024} \\
    & & Wrapper & Reinforcement Learning & Generative-methods~\citep{hsu2024safety, prashant2025guaranteeing,gardille}\\ \midrule
\rowcolor{Gray} Run-time Operations & Detection of Anomalies & Wrapper & Semi-Supervised & Formal Verification~\citep{stahl2021} , Gaussian mixture models~\citep{schnell2020}, Generative Models~\citep{isaku2025digital} , Foundation Models~\citep{greer2024towards}, Vision Transformer~ \citep{dosovitskiy2020image} \\
& & Filter & Semi-supervised & Normalising flows~\citep{Feng2023NF}  \\ \cmidrule(l){2-5}
 & Detecting data shifts (Covariate) & Wrapper & Unsupervised & Foundation Models~ \citep{heng2025detecting} \\ 
 \rowcolor{Gray}&  & Wrapper & Semi-Supervised & Statistical~\citep{filos2020} , Temperature scaling~ \citep{hsu2020generalized}  \\ \cmidrule(l){2-5}
 & Detecting data shifts (Semantic) & Wrapper & Supervised & Statistical~\citep{zamzmi2024out} \\ 
 \rowcolor{Gray}& & Wrapper & Semi-Supervised & Foundation models + generative models~\citep{Cultrera2023}, \\ \cmidrule(l){2-5}
 & Detecting Novel Scenarios & Wrapper & Semi-Supervised & Uncertainty measures~\citep{mcallister2019}, Generative models~\citep{zheng2025diffusionbased}, Foundation models~\citep{ganai2025real} \\ 
 \rowcolor{Gray}&  & Wrapper & Reinforcement Learning & Switching logic~\citep{musau2022using} \\ \midrule
Other Techniques & - & Wrapper & Unsupervised & Distance-based~\citep{goyal2020,denouden2018improving,jungo2023unsupervised,sun22d,sikar2024evaluationautonomoussystemsdata}, Density-based~\citep{ranjbar2022,ramakrishna2022efficient,zong2018deep} \\
 \rowcolor{Gray} & - & Wrapper & Semi-supervised & Ensemble~ \citep{Kirchheim_2024_WACV}, Foundation models~\citep{fort2021exploring}  \\
& - & Wrapper & Supervised with OSR &  Semantic segmentation~\citep{shoeb2025out}\\ \bottomrule

\end{longtable}

%\end{landscape}

\section{Discussion}\label{sec:discussion}

There has been significant progress in the effective implementation of OOD detection in autonomous systems recently. However, there are a number of considerations to ensure safety and reliability when designing and integrating OOD detection components into the system architecture.

We note that there is a tendency to rely on  metrics as a guarantee of safety, rather than taking a system centric view of safety. Indeed, the recall or accuracy of ML models does not directly equate to their safety level. We have noticed that some authors in the literature analyse the recall or accuracy of their OOD models to determine how safely they will operate; they conflate safe/unsafe with ID/OoD. Even performing correctly 99\% of the time does not guarantee safety. If a system fails 1\% of the time during operation then that is sufficient to lead to serious hazards and highlights the importance of redundant techniques for monitoring these algorithms at run-time~\citep{Lohn2020EstimatingTB}, including OOD monitoring. More frequently a trade-off between multiple metrics is required to measure the OOD detector's performance. This trade-off needs to be considered and analysed through the lens of safety~\citep{henriksson2019metrics,Tambon2021HowTC}.

We also note that, paradoxically, just because a sample is identified as being OOD, this does not mean that it is unsafe. Autonomous systems can encounter unknown scenarios and operate safely~\citep{mcallister2019}. In fact, a data distribution shift may be beneficial, for example, it could over-represent a minority group in the training data of a medical autonomous system~\citep{petersen2022}. Since our safety case has been created with an assumption of an intended operation domain, however, we cannot guarantee its safety in these OOD scenarios.

Moreover, in the \textit{System Specification} phase  in Fig.~\ref{fig:ML-Process-Diagram}, the safety assurance arguments, context and justifications can only be specified as ``intended functionality'' with a set of high-level goals and objectives or by iteratively refining the safety assurance. This can lead to a \textbf{semantic gap} \citep{burton2020MindTheGaps,Perez-Cerrolaza2024} between the intended functionality and the specified functionality. Producing a correct, accurate and complete representation of the ``intended functionality'' is a challenge and this increases the likelihood of OOD data at run-time. Additionally, the system specification may be amended after deployment (e.g., if we use continuous learning in an ML-based OOD detector which updates the model and consequently its specification) and this further increases the semantic gap.

This semantic gap must also be offset against the \textbf{feature gap}. Even if there is a large semantic difference between ID and OOD data, there may not be a corresponding difference between the underlying data features. For example, an ID and an OOD object may have similar texture or shape. This makes them difficult to distinguish for algorithms that learn data features. This highlights the need for multi-faceted OOD detection and monitoring for autonomous systems to assure their safety.

The ML component's performance and its impact on the \textbf{system} must be analysed, evidenced and argued for the system to be considered safe~\citep{sinha2023systemlevelviewoutofdistributiondata}. Most OOD research focuses on OOD data's impact on the ML component rather than how its impacts propagate and impact the autonomous system. \citet{ramakrishna2022dynamic} notes that safety cases use general patterns to argue system safety and that these general patterns  do not capture how the assurance of the ML component impacts the overall assurance of the system nor are assumptions regarding the trained ML component captured. Another challenge is ``\textit{to estimate the frequency and extent of departures from designed conditions that the AI
will encounter in the real world and estimate the AI’s performance in those conditions}'' \citep{Lohn2020EstimatingTB}, i.e., ``how will it perform outside the ODD?'' One solution is specified run-time robustness constraints which limit the autonomous system's operational domain to that specified in the \ODD description but this constrains the autonomous system's applicability and prevents generalisation to new scenarios and domains. 

A further trade-off is the sensitivity-robustness dilemma \citep{zhang2024the} which offsets fine-grained accuracy in detecting subtle semantic shifts (for example detecting near-OoD data categories) against generalisability and robustness to noise or covariate shifts. The classification accuracy of algorithms such as those used in foundation models is susceptible to even slight noise or minor domain shifts and can catastrophically collapse. This trade-off compromises the trustworthiness of the model and significantly limits its practical deployment in dynamic real-world environments. Hence, a truly robust OOD detection algorithm for autonomous systems must address this trade-off.

% Not sure what this is trying to say
%Many safety standards use the concept of system safety integrity levels (SIL), also called design assurance levels (DAL). These state the relative level of risk-reduction and reliability requirements need to achieve a particular level. We note, however that these levels vary across industries in terms of magnitude, labelling, intent, and implementation as noted by~\citet{Lohn2020EstimatingTB}. These levels are defined in terms of system level functionality, the ML components do not necessarily need to achieve a particular level. Therefore, for a low SIL, the autonomous system may switch to a safe fallback mode if the ML component has low confidence in its inference and thus its  ability to perform safely. Current \OOD methods can safely monitor the input data, the ML component and its inferences and switch. However, for higher SILs, current \OOD techniques may not provide sufficient confidence that the system fulfils these more demanding safety requirements across the \ODD.

\section{Conclusions, Challenges and Future Work}\label{sec:conclusion}
Proving acceptable safety through informed design and simple act/fail/fix development is  very difficult and a long-standing challenge for autonomous systems~\citep{koopman2018heavy,ranjbar2022}. These challenges as the level of autonomy increases. Due to the complexity and dynamism of the real-world, autonomous systems will encounter OOD data in both their train and test data and, will encounter unknown (OOD) data and scenarios at run-time. To operate safely these systems should be able to, at the very least, identify and ideally recover from such scenarios, without catastrophic failure~\citep{filos2020}. 
Thus, OOD detection will be a vital component of a ``safety-cage architecture''~\citep{henriksson2019} to assure and ensure the safety of autonomous systems as they become ubiquitous. 

In this paper, we considered an ML system lifecycle, and  showed how a range of OOD detection algorithms could be used throughout the lifecycle and across the system architecture to aid in the development, and assurance, of safety critical autonomous systems. Whilst there has been a great deal of work on developing OOD techniques, significant challenges still exist before we can guarantee autonomous systems safety in open-world environments.

%There remain a number of challenges for \OOD detection (particularly \OOD for the safety assurance of autonomous systems) which identify how researchers and developers can build on our proposed lifecycle. They highlight the requirements for future work to ultimately produce inherently safe design (safety strategy 1) for autonomous systems operating in safety-critical domains.

\subsection{Data}\label{sec:data}

%Data engineering for \OOD analyses has many challenges but there have only been limited research contributions. For safety assurance, to develop the safety case and argue safety compliance, the data must provide a complete, relevant, balanced and accurate representation of the intended functionalities \citep{ashmore2021assuring} and the autonomous system must have the ability to identify unknowns~\citep{koopman2018heavy,schleiss2022}. 
The size and complexity of the data distributions in open-world contexts mean that it cannot be fully specified at design time nor tested exhaustively during verification. The long tail of unseen scenarios~\citep{koopman2018heavy} renders probabilistic assessment unlikely to identify good coverage and  it can be expensive and dangerous to collect OOD examples (for example OOD scenarios that lead to serious accidents).  \citet{neto2022} identify the need to establish a cost-effective method for accurate labelling of training, validation and testing datasets for supervised OOD modelling.  Additionally,  data and scenarios may be within the \ODD but unsafe due to incomplete specification of the functionality, gaps in the training data or incomplete learning by the ML model.  This semantic gap~\citep{burton2020MindTheGaps,Perez-Cerrolaza2024} between the functionality specified at design and development time and the intended functionality at run-time increases the likelihood of OOD data at run-time.  

%Data augmentation and large-scale simulation are currently unlikely to capture all scenarios~\citep{ding2023} and simulation is not always faithful~\citep{ryan2024safety}.  In particular, t
There is a scarcity of data related to safety-critical edge cases~\citep{neto2022}. Edge cases may have only small input perturbations compared to ID data and this can lead an ML model to recommend a potentially unsafe action. Future research will define ways to identify representative edge cases and benchmarks for how ML models take them into account~\citep{neto2022}. Generative models such as diffusion models are starting to show promise due to their ability to model complex data distributions and to generate high-quality synthetic data~\citep{lyhs2025bootstrapping}

Closing the performance gap between detecting far-OOD and near-OOD data is an open challenge for OOD detection~\citep{fort2021exploring,Lohn2020EstimatingTB,mohseni2022}, particularly where the data are ``\textit{visually similar to IID data but yet outliers w.r.t. semantic meanings}''~\citep{mohseni2022}. This is a common challenge for monitoring fine-grained image analysis algorithms. This also extends to certain inputs that fall inside the \ODD but are conceptually harder to identify by OOD methods. \citet{schwaiger2020uncertainty} state that more research is required to investigate the phenomenon. At present, if weakly covered distribution areas are identified during OOD detection, safety engineers modify the ML model's functional scope by refining the requirements or reducing the scope of the ODD~\citep{henriksson2023}. 

\subsection{Benchmark Methodologies and Baselines}\label{sec:benchmarks}
The systematic error analysis of ML model training is vital for developing safe models and to make safety guarantees, but limited research addresses the challenge of systematicity~\citep{Perez-Cerrolaza2024}. We need a full understanding of how OOD failures in ML components impact system safety. There has only been limited work on verifying OOD detection approaches in safety-critical domains \citep{Tambon2021HowTC} leaving a requirement for assessment frameworks for benchmarking and establishing baselines. \citet{hendrycks22a_MLS} provide a framework of algorithms and \citet{yang2022openood} provide data sets, both cover image analyses. However, standard baseline models, methodologies or data sets to compare against are needed for other domains as algorithm performance depends heavily on the choice of OOD samples used for evaluation.
Furthermore, for many ML algorithms such as neural networks, their performance comes with no theoretical guarantees which are critical for safety certification~\citep{Tambon2021HowTC} so benchmarks and baselines are increasingly important.

%More research is needed to build a classification scheme of \OOD algorithms and to understand the relationship between various types of algorithms and how to mitigate underlying uncertainty. System designers need to be able to identify the functional uncertainty in the underlying system architecture. Autonomous systems need to be able to detect when they are encountering \OOD scenarios across the \ODD (and potentially beyond) to remain safe in uncertain environments. 

%\subsection{Medical Domain}
%Detecting \OOD throughout the system lifecycle is a critical challenge for medical ML but the safety risks have prevented widespread adoption up to now. For example, \OOD detection on 3D medical images has seen limited research,  even though ``\textit{3D medical image segmentation is one of the most addressed tasks in medical imaging}''~\citep{Vasiliuk2023}. To assure the safety of an autonomous system in the medical domain, $D_{train}$ must meet the four data desiderata for ML systems, and we must be able to provide a complete, correct and representative specification of the intended safety functionalities of the system. Current models trained on small unrepresentative datasets are not robust against differing recording environments or devices, patient demographics, hospital types, healthcare systems, or healthcare policy shifts~\citep{petersen2022}.  We need standardised datasets, baseline models, benchmark methodologies, and standardised metrics (discussed next) to enable comparisons and to provide explanations for decisions.

\subsection{Metrics}
A metrics standard which defines requirements and methods for the verification and validation of \OOD for autonomous systems is needed to ensure the safe and resilient handling of OOD scenarios for autonomous systems~\citep{greer2024towards}. Standard metrics must cover a number of properties of the data, ML models and autonomous system. \citet{Perez-Cerrolaza2024} posit the important properties of safety-critical systems are: auditability; data quality; explainability/interpretability; monitorability; provability; and robustness. Data quality covers relevance; completeness; balance; and accuracy~\citep{ashmore2021assuring} which are related to the system's  application and domain. Model quality analyses \citep{rainio2024evaluation} show the importance of metrics and their correct usage to prevent false conclusions and untrustworthy analyses. 
% The ability to identify when the ML model is likely mistaken is also still a missing but safety-critical prerequisite for the usage of ML in autonomous systems~\citep{bruggemann2020}.
 
Since single metrics are unlikely to be sufficient, a systematic approach for the combination of metrics is also needed. Such an approach should be analysed through the lens of safety~\citep{Tambon2021HowTC}. For example, the best trade-off for safety may not necessarily be the one with the highest model recall. For a safety-critical application, a low value of false negatives (OOD classed as ID) would be more important than a high area under the receiver operator curve (AUROC \citep{bradley1997}) value though too many false positives (ID classified as OOD) would make the system too slow~\citep{henriksson2019,Tambon2021HowTC} and introduce operator distrust. Moreover, treating false positives or false negatives equally can also be misleading so work is needed to develop a more informative metric  with task and situational awareness rather than generalising to the whole distribution (see section \ref{sec:task_specif}). \citet{sinha2023systemlevelviewoutofdistributiondata} note that an ideal OOD detector should distinguish
consistent model errors or distribution shifts from sporadic errors caused by OOD instances. An ideal OOD metric should also differentiate between near-OOD semantic shifts that introduce novel classes and far-OOD covariate shifts (anomalies in the data) \citep{yang2024}.

\subsection{Time Horizon}

We note that many autonomous systems have to operate over different time horizons. An autonomous robot must take actions and react to changing environments in real-time as it navigates. This necessitates OOD monitoring for real-time analysis and decision making. The robot will have to process heterogeneous data from a variety of sources (sensors, the environment, etc.) almost simultaneously. Handling this volume and variety of data is a major challenge. Also, the more complex the OOD algorithm, the higher the likelihood of decision latency \citep{baccari2024}. 
By contrast an autonomous system for longer-term tasks such as route planning, will require the OOD monitor to analyse episodic interactions and time correlations should be accounted for~\citep{sinha2023systemlevelviewoutofdistributiondata}.

\subsection{Task Specificity}\label{sec:task_specif}
An ideal perception system will detect when task-salient features of the data are unfamiliar or uncertain, while ignoring task-irrelevant features. \citet{mcallister2019} give the example of a ground robot that performs collision avoidance using a model trained on camera data from buildings with white ceilings. The robot should function normally in buildings with black ceilings but DNN-based approaches often have high uncertainty in these circumstances, even though the ceiling colour is irrelevant to the task. Generalised OOD detection across all data features is overly pessimistic as it is unaware of which data features are relevant to the robot’s task or safety. Hence, 
\citet{mcallister2019,henriksson2023} propose an ideal OOD detector is task-aware and can ignore task-irrelevant features. 
Furthermore, \citet{novello2024out} separate task-based OOD and task-agnostic OOD. Task-based is supervised learning where the OOD detector classifies data and inferences into a set of classes (tasks) including an OOD class and task-agnostic OOD is supervised or semi-supervised learning where IID is inside the class and everything else is OOD. 
 Incorporating context, salience and task-awareness are important challenges for OOD detection and the metrics used for determining OOD, so we can assure and argue the autonomous system's safety.

\subsection{Computational Complexity and Scalability} % https://deepgram.com/ai-glossary/out-of-distribution-detection
%\citep{hendrycks22a_MLS}  
Most algorithms described in the literature are evaluated on small or medium-sized data sets~\citep{baccari2024}. Sensor data collected by autonomous systems are usually large and heterogeneous.  Hence, collecting high-quality, representative datasets is a huge challenge and even then they may not cover the full distribution. For an autonomous car, thousands of hours of real-world driving data are required, both from manually driven and autonomous vehicles, in multiple locations, times and weathers to enable robust results. Coupled with this, an autonomous robot navigates at high speed in high-density and diverse environments so must make rapid decisions which requires very rapid processing to guarantee safety requirements are met. The higher the complexity of the OOD algorithm, the greater the risk of associated latency~\citep{baccari2024}. Furthermore, computational resources may be constrained particularly on smaller robots so this needs to be carefully considered within the system and safety requirements. A reliable OOD detector must accommodate these constraints so it can guarantee fast and safe OOD detection (worst-case execution time guarantees) in a variety of environments. Very few papers in the literature provide run-time calculations and even fewer consider worst-case execution times. For safety assurance of autonomous systems, the data management and processing has many challenges with only limited research contributions~\citep{Perez-Cerrolaza2024}.

\subsection{Explainability}
Explainability makes OOD monitoring transparent by providing human-interpretable explanations. LLMs can produce explanations for OOD detection in safety-critical systems where transparency and interpretability are important for trust \citep{xu2024large}. It will allow us to transition from simply detecting OOD to understanding and managing OOD. It is an active area of research as it increases trust in the system's decisions and provides evidence for safety assurance including detecting when
safety requirements and assumptions are violated~\citep{jia2022explain,tekkesinoglu2025}. However, explanations need to be carefully considered to ``meet the diverse needs of stakeholders'', to be relevant to the task and at the required level of autonomy of the system \citep{tekkesinoglu2025}. As systems move to higher levels of autonomy, the breadth and complexity of OOD data encountered will increase so the capabilities of the OOD detection and explanation will need to increase accordingly, throughout the lifecycle 

Explainers can be local (per datum) or global (overall model behaviour) and model-specific (filters) or model-agnostic (wrappers) so align with the OOD paradigm described in the paper. They can analyse the training and test data distributions, and the model. Additional data may then be collected to fill any data gaps during the \textit{System Specification}, \textit{Data Management} and \textit{Model Learning \& Verification} stages making the deployed system more resilient and reducing model errors.  The explanation can allow the model to be refined to reduce model errors.

\subsection{Foundation Models (FMs)}
Foundation models including LLMs and multi-modal models are a new paradigm for OOD and their safety assurance is a nascent but expanding research area. They can be used throughout the lifecycle starting with \textit{Data Management}. FMs can also map heterogeneous data sources to a single latent space (embeddings) to provide representation, task, embodiment and environmental agnosticism for robots. At the moment this work is limited to supervised learning \citep{wang2024hpt}. They can also synthetically generate data to augment the datasets for training both $M$ and $M_{OOD}$, see for example \citep{chen2024fodfom}. However, the synthetic data generation is bootstrapped using the existing dataset. If we need to augment the data then we must have identified gaps and deficiencies.  Hence, using FMs can further reinforce biases or extend noise already present in the data. 
While FM data management and augmentation offers promise, to ensure robustness of the models and assure safety, rigorous validation and bias mitigation strategies are needed.

Foundation models underpin agentic AI \citep{Acharya2025agentic} and this new discipline is moving from traditional narrow AI development trained on small task-specific data sets to a more integrated, adaptable and transferable approach. Thus a FM-based OOD detector will have a broader, more generalised capability than existing OOD detectors which are bespoke to the ML component (such as a pedestrian detector). An FM can then be efficiently fine-tuned for specific contexts. This has the potential to introduce more consistent and scalable safety mechanisms that can be transferred across applications (for example transfer learning to  generalise a robot to novel skills~\citep{brohan2023rt}. However, by using transfer learning for $M_{OOD}$, there may be information leakage \citep{li2025recent}. FMs are trained on vast datasets scraped from heterogenous sources. During training, they may have already encountered OOD data points
or OOD features. The model's OOD detection capability will be compromised. This necessitates carefully curated training data (see  section \ref{sec:data}) with benchmarks and standards (see section \ref{sec:benchmarks}).

Authors have also found that the process of fine-tuning foundation models with task-specific data may lead to reduced robustness of out-of-distribution (OOD) detection models \citep{andreassen2021evolution}. Standard fine-tuning approaches may lose ID knowledge embedded in the FM reducing its ability to generalise. This issue is closely related to the "sensitive-robust" dilemma when there is an inherent trade-off between performance and robustness \citep{andreassen2021evolution}. New fine-tuning approaches are needed and to identify ``a single set of hyper-parameters that works effectively across multiple fine-tuning
distributions''\citep{choi2024autoft}.

Researchers are exploring a number of directions and applications for FMs but there is still much work to do assuring their safety and achieving safety guarantees. \citet{hsu2024safety} observe that ``\textit{while traditional model-based safe control methods struggle with generalizability and scalability, emerging data-driven approaches (such as deep neural networks) tend to lack well-understood guarantees, which can result in unpredictable catastrophic failures}''.  \citet{hsu2024safety,ranjbar2022} propose that a combination of model-based and data-driven techniques will be required to assure the safety of autonomous systems in the future.

\section*{Statements and Declarations}
\subsection*{Acknowledgements}
The work was supported by the Centre for Assuring Autonomy (CfAA), a partnership between Lloyd’s Register Foundation and the University of York (\url{https://www.york.ac.uk/assuring-autonomy/}).
\subsection*{Competing Interests}
The authors have no relevant financial or non-financial interests to disclose.
\subsection*{Data Availability}
No datasets were generated or analysed during this work.

%\include{Todo}

%Bibliography
\bibliographystyle{abbrvnat}  
\bibliography{main}

\end{document}